\documentclass[letterpaper]{article} 
\usepackage{aaai2026}  
\usepackage{times}  
\usepackage{helvet}  
\usepackage{courier}  
\usepackage[hyphens]{url}  
\usepackage{graphicx} 
\usepackage{natbib}  
\usepackage{caption} 
\usepackage{algorithm}
\usepackage{algorithmic}
\usepackage{subfigure}

\usepackage{booktabs}
\usepackage{amsfonts}
\usepackage{multirow}
\usepackage{amsmath}
\usepackage{amsthm}
\graphicspath{{./figures/}}
\theoremstyle{plain}
\newtheorem{theorem}{Theorem}[section]
\newtheorem{proposition}[theorem]{Proposition}
\newtheorem{lemma}[theorem]{Lemma}

\theoremstyle{definition}
\newtheorem{definition}[theorem]{Definition}
\newtheorem{assumption}[theorem]{Assumption}
\theoremstyle{remark}
\newtheorem{remark}[theorem]{Remark}
\usepackage{multirow} 
\usepackage{tablefootnote}
\newenvironment{proofsketch}{%
  \begin{proof}%
}{%
  \end{proof}%
}

\usepackage{newfloat}
\usepackage{listings}
\DeclareCaptionStyle{ruled}{labelfont=normalfont,labelsep=colon,strut=off} 
\floatstyle{ruled}
\newfloat{listing}{tb}{lst}{}
\floatname{listing}{Listing}
\title{An Improved Privacy and Utility Analysis of Differentially Private SGD \\ with Bounded Domain and Smooth Losses}
\author{
    Hao Liang\textsuperscript{\rm 1}, Wanrong Zhang\textsuperscript{\rm 2}, Xinlei He\textsuperscript{\rm 1}, Kaishun Wu\textsuperscript{\rm 1}, Hong Xing\textsuperscript{\rm 1,3}\thanks{Corresponding author: Hong Xing (hongxing@ust.hk).}\\
}
\affiliations{
    \textsuperscript{\rm 1}Information Hub, The Hong Kong University of Science and Technology (Guangzhou)\\
    \textsuperscript{\rm 2}Harvard University\\
    \textsuperscript{\rm 3}Department of ECE, The Hong Kong University of Science and Technology\\

    hliang346@connect.hkust-gz.edu.cn,  wanrongzhang@fas.harvard.edu, \{xinleihe, wuks\}@hkust-gz.edu.cn, hongxing@ust.hk
}

\usepackage{bibentry}

\begin{document}

\maketitle

\begin{abstract}
Differentially Private Stochastic Gradient Descent (DPSGD) is widely used to protect sensitive data during the training of machine learning models, but its privacy guarantee often comes at a large cost of model performance due to the lack of tight theoretical bounds quantifying privacy loss. While recent efforts have achieved more accurate privacy guarantees, they still impose some assumptions prohibited from practical applications, such as convexity and complex parameter requirements, and rarely investigate in-depth the impact of privacy mechanisms on the model's utility. In this paper, we provide a rigorous privacy characterization for DPSGD with general L-smooth and non-convex loss functions, revealing converged privacy loss with iteration in bounded-domain cases. Specifically, we track the privacy loss over multiple iterations, leveraging the noisy smooth-reduction property, and further establish comprehensive convergence analysis in different scenarios. In particular, we show that for DPSGD with a bounded domain, (i) the privacy loss can still converge without the convexity assumption, (ii) a smaller bounded diameter can improve both privacy and utility simultaneously under certain conditions, and (iii) the attainable big-O order of the privacy utility trade-off for DPSGD with gradient clipping (DPSGD-GC) and for DPSGD-GC with bounded domain (DPSGD-DC) and strongly convex population risk function, respectively. Experiments via membership inference attack (MIA) in a practical setting validate insights gained from the theoretical results.
\end{abstract}

\begin{links}
    \link{Code}{https://github.com/HauLiang/DPSGD-DC}
    \link{Extended version}{https://arxiv.org/abs/2502.17772}
\end{links}

\section{Introduction}
\label{s1}

Differentially Private Stochastic Gradient Descent (DPSGD) \cite{abadi2016deep} has emerged as the leading defense mechanism to protect personal sensitive data in training of machine learning models. However, achieving good performance with DPSGD often comes with a significant privacy cost. A fundamental question, therefore, is how to precisely quantify the privacy loss associated with DPSGD. 

Previous methods for quantifying privacy loss include strong composition \cite{dwork2010boosting,bassily2014private,kairouz2015composition}, moments accountant \cite{abadi2016deep}, R\'enyi Differential Privacy (RDP) \cite{mironov2017renyi,mironov2019r}, and Gaussian Differential Privacy (GDP) \cite{dong2022gaussian}, along with several numerical composition methods \cite{koskela2020computing,gopi2021numerical}. These methods primarily rely on composition theorems, assuming that all intermediate models are revealed during the training procedure, which leads to an overestimation of privacy loss. While numerical composition methods aim to tightly characterize the privacy loss, they still operate under this same assumption.

To address this overestimation, recent works have focused solely on the privacy guarantees of the final output. For instance, the privacy amplification by iteration \cite{feldman2018privacy} demonstrated that withholding intermediate results significantly enhances privacy guarantees for smooth and convex objectives.  Building upon this, \citet{chourasia2021differential} suggest that the privacy loss of DPGD, the full batch version of DPSGD, may converge exponentially fast for smooth and strongly convex objectives. Furthermore, results by \citet{ye2022differentially} as well as \citet{ryffel2022differential} extended this analysis to assess the privacy loss of DPSGD, although both studies rely on the assumption of strong convexity.

More recently, the work by \citet{altschuler2022privacy} and its extension \cite{altschuler2024privacy} established a constant upper bound on privacy loss after a burn-in period for Lipschitz continuous and smooth convex losses over a bounded domain. However, this analytical result is limited by its reliance on the convexity assumption and strict restrictions on the Rényi parameter $\alpha$, which hinders its broader applicability. In order to relax several strong assumptions, \citet{kong2024privacy} provided an analysis of weakly-convex smooth losses in the case where data is traversed cyclically. Later, \citet{chien2024convergent} suggest precisely tracking the privacy leakage incurred before reaching the constant upper bound by solving an optimization problem. However, this result is formulated as a complex optimization problem rather than a closed-form expression, making it hard to operationalize. Notably, most recent methods necessitate double clipping of both gradients and parameters due to the additional bounded domain assumption. These methods, however, do not provide a thorough utility analysis or experimental results, leaving their practical performance and trade-offs underexplored.

We outline the main contributions of this paper below and provide a comparison of the key assumptions and theoretical results with the most relevant works in Table~\ref{table:compare}.

\begin{table*}
  \caption{Comparison of the $(\alpha, \varepsilon)$-RDP guarantee and assumptions needed by different works for DPSGD, where $b$ is the batch size, $n$ is the dataset size, $\eta$ is the step size, $C$ is the gradient clipping norm bound, $D$ is the diameter of the parameter domain, $\sigma_{\text{DP}}$ is the noise scale, and $T$ is the number of iterations. ``$\ddag$" is exclusively suitable for cyclic data traversal cases. ``$\dag$" indicates that a tighter bound can be obtained under additional assumptions on the R\'enyi parameters. $\alpha^*(q,\sigma)$ is defined as the largest $\alpha$ that satisfies both $\alpha \leq K \sigma^2/2-2\log \sigma$ and $\alpha \leq\left(K^2 \sigma^2 / 2-\log 5-2\log \sigma\right) /\left(K+\log (q \alpha)+1 /(2 \sigma^2)\right)$, where $K=\log(1+1/(q(\alpha-1)))$.}
  \label{table:compare}
  \centering
  \resizebox{\textwidth}{!}{
  \begin{tabular}{lccccr}
    \toprule
    Reference & Assumptions  & Domain & Privacy Guarantee & Utility Analysis? \\
    \midrule
    \multirow{2}{*}{\citet{feldman2018privacy}}   & \multirow{2}{*}{convex, $L$-smooth, $M$-Lipschitz, $\eta\leq 2/L$} & \multirow{2}{*}{unbounded} & \multirow{2}{*}{$\mathcal{O}\left( \frac{ \alpha M^2}{b^2 \sigma_{\text{DP}}^2} T\right)$} & \multirow{2}{*}{$\surd$} \\
    \\ \midrule 
    \multirow{2}{*}{\citet{altschuler2022privacy}}   & convex, $L$-smooth, $M$-Lipschitz, $\eta\leq 2/L$ & \multirow{2}{*}{bounded} & \multirow{2}{*}{$\mathcal{O}\left( \frac{\alpha M^2}{n^2\sigma_{\text{DP}}^2} \min \left\{T, \frac{Dn}{\eta M}\right\}\right)$} & \multirow{2}{*}{$\times$}\\
    & $b\leq n/5$, $\sigma_{\text{DP}}>8\sqrt{2}M/b$, $\alpha\leq \alpha^*(b/n,\frac{b\sigma_{\text{DP}}}{2\sqrt{2}M})$  & \\ \midrule
    \multirow{2}{*}{\citet{kong2024privacy}$^\ddag$}      & \multirow{2}{*}{$m$-weakly convex, $L$-smooth, $\eta\leq\frac{1}{2(m+L)}$} & \multirow{2}{*}{bounded} &  \multirow{2}{*}{$\mathcal{O}\left(\frac{\alpha}{\eta^2\sigma_{\text{DP}}^2}(D\sqrt{1+2\eta m[1+\frac{m}{2(L+m)}]}+\frac{\eta C}{b})^2\right)$}  & \multirow{2}{*}{$\times$}      \\
       \\ \midrule
    \multirow{2}{*}{\citet{chien2024convergent}}      & $M$-Lipschitz & unbounded &  \multirow{2}{*}{w/o analytical form}  & \multirow{2}{*}{$\times$}      \\
     & $L$-smooth or $(L,\lambda)$-H\"older continuous gradient & /bounded\\ \midrule
    \textbf{Ours}$^\dag$ & $L$-smooth & unbounded & $\mathcal{O}\left( \frac{ \alpha C^2}{n b\sigma_{\text{DP}}^2} T\right)$& $\surd$ \\[5pt]
    \textbf{Ours}$^\dag$ & $L$-smooth & bounded & $\mathcal{O}\left(\frac{\alpha C^2}{nb\sigma_{\textup{DP}}^2}\min\left\{T, \frac{(1+\eta L)^2nbD^2}{\eta^2C^2}\right\}\right)$
    & $\surd$ \\
    \bottomrule
  \end{tabular}
  }
\end{table*}

\subsection{Contributions}

In this paper, we present a precise analytical 
characterization of privacy bounds for DPSGD that focus on smooth loss functions without relying on convexity assumptions or restrictive R\'enyi parameter conditions. Our general results encompass DPSGD applied to both unbounded and bounded domains. Additionally, we establish utility guarantees based on the derived RDP bounds, offering an intuitive perspective on privacy-utility trade-offs. To demonstrate the practicality and validity of our theoretical findings, we conduct extensive numerical simulations, which confirm the effectiveness and rationality of the proposed bounds. Our contributions are as follows:
\begin{itemize}
\item We analyze the noisy smooth-reduction behavior of the shifted R\'enyi divergence for smooth objectives. This analysis enables the derivation of closed-form RDP guarantees for DPSGD applied to both unbounded and bounded domains.

\item We establish the convergence behavior for DPSGD with smooth loss functions in unbounded domains and strongly convex smooth loss functions in bounded domains. 
Our results provide the privacy-utility trade-offs under the computed RDP bounds.
\item To validate these theoretical findings, we examine the privacy parameters estimated by the membership inference attack (MIA). Extensive experiments illustrate the effectiveness and rationality of the proposed bounds.
\end{itemize}

To clearly illustrate the effectiveness of our proposed privacy bound, we here provide detailed comparisons with several prominent existing approaches, including: a) \citet{feldman2018privacy}, (b) the combined analysis by \citet{mironov2017renyi} and \citet{mironov2019r}, c) \citet{altschuler2022privacy}, and d) \citet{kong2024privacy}. The detailed setting can be found in Appendix~\ref{app:num-set}. As shown in Figure~\ref{fig:compare-theoretical}, our analysis demonstrates strict improvement on all existing privacy bounds, except for \citet{altschuler2022privacy}. The reason for which their bound appeared tighter is their additional assumptions—including convexity and more restrictive conditions (as summarized in Table~\ref{table:compare})—while our analysis relies only on the smoothness of the loss function and far weaker assumptions.

\begin{figure}[!t]
\centering
\includegraphics[width=0.46\textwidth]{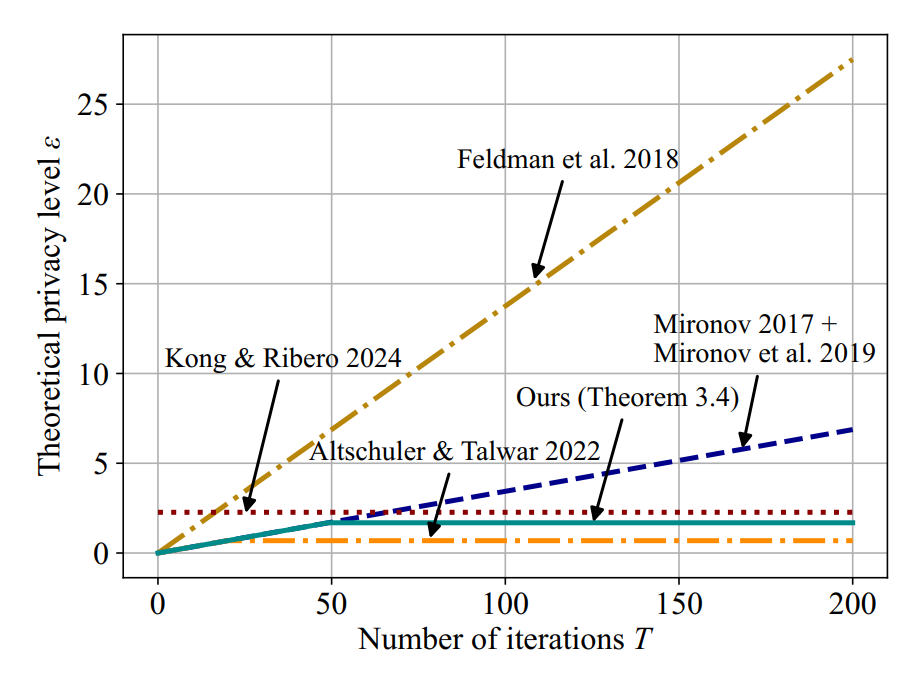}
\caption{Comparison of our theoretical $(\alpha, \varepsilon)$-RDP bound for DPSGD-DC with existing approaches. Detailed assumptions required by each method have been summarized in Table~\ref{table:compare}.}
\label{fig:compare-theoretical}
\end{figure}

\subsection{Other Related Works}

In addition to privacy analysis, the utility (convergence) of private optimization algorithms has been extensively studied. This line of works typically focus on understanding how the number of iterations affects the convergence behavior of the algorithm. Below, we provide a brief review of utility analysis for DPSGD.

In 2014, \citet{bassily2014private} analyzed the optimal utility guarantees of DPSGD under the assumption of Lipschitz continuity, considering both convex and strongly convex cases. Then, based on the additional assumption of the gradient distribution, \citet{chen2020understanding} studied the convergence of DPSGD with gradient clipping (DPSGD-GC) and derived a utility bound for the non-convex setting. The work by \citet{song2021evading} explored the convergence of DPSGD-GC for generalized linear models, noting that, in the worst case, the utility can remain constant relative to the original objective. Later, \citet{fang2023improved} further refined this analysis for smooth and unconstrained problems, providing more precise convergence results. However,  many of these studies fix a specific value for the clipping threshold $C$, which may be adjusted due to privacy requirements. 

More recently, \citet{koloskova2023revisiting} characterized the convergence guarantees for DPSGD-GC across various clipping thresholds $C$ in the non-convex setting. While this work provides valuable convergence insights for DPSGD-GC, recent privacy characterizations have introduced the need for double clipping—clipping both gradients and parameters—due to the additional assumption of bounded domains. The convergence analysis involving double clipping has not been thoroughly explored in the existing literature.

\subsection{Organization}
The rest of this paper is organized as follows: In the next section, we recall the relevant preliminaries. Our main results are presented in Section \ref{s3}. Numerical results are provided in Section \ref{s4}. Finally, Section \ref{s5} concludes with a discussion of future research directions motivated by our findings. Proof details are deferred to Appendices.

\section{Preliminaries}
\label{s2}

In this section, we introduce the foundational concepts and definitions relevant to our analysis. We start with our notations, which will be used throughout this paper.

\textbf{Notations.} Let $\operatorname{Pr}[\cdot]$ denote the probability of a random event, and $\mathbb{P}_{\boldsymbol{\mu}}$ be the law of a random variable $\boldsymbol{\mu}$. We refer to two datasets $\mathcal{D}$ and $\mathcal{D}^\prime$ as \emph{adjacent} if they differ from each other by adding or removing only one data point.

\subsection{R\'enyi Differential Privacy (RDP)}

We first recall the formal definition of differential privacy (DP) and RDP.

\begin{definition}
\textup{(Differential privacy \cite{dwork2006our}).}
For $\epsilon\geq 0$, $\delta\in[0,1]$, a randomized mechanism $\mathcal{M}:\mathcal{X}\mapsto \mathcal{Y}$ is \emph{$(\epsilon,\delta)$-DP} if, for every pair of adjacent datasets, $\mathcal{D}, \mathcal{D}^\prime \subseteq \mathcal{X}$, and for any subset of outputs $\mathcal{S}\subseteq \mathcal{Y}$, we have
\begin{equation}
    \operatorname{Pr}[\mathcal{M}(\mathcal{D})\in \mathcal{S}]\leq \exp(\epsilon)\operatorname{Pr}[\mathcal{M}(\mathcal{D}^\prime)\in \mathcal{S}]+\delta.
\end{equation}
\end{definition}

Throughout this paper, we use RDP, a more efficient approach for tracking privacy loss than DP, as our primary framework for privacy analysis. RDP provides a relaxation of DP based on \emph{R\'enyi divergence}, which is defined as follows.

\begin{definition}
\textup{(R\'enyi divergence \cite{renyi1961measures}).} For adjacent datasets $\mathcal{D}$ and $\mathcal{D}^\prime$, a randomized mechanism $\mathcal{M}:\mathcal{X}\mapsto \mathcal{Y}$, and an outcome $s\in \mathcal{Y}$, the R\'enyi divergence of a finite order $\alpha\neq1$ between $\mathcal{M}(\mathcal{D})$ and $\mathcal{M}(\mathcal{D}^\prime)$ is defined as
\begin{equation}
\begin{aligned}
&D_\alpha(\mathbb{P}_{\mathcal{M}(\mathcal{D})}||\mathbb{P}_{\mathcal{M}(\mathcal{D}^\prime)}) \\
&=\frac{1}{\alpha-1} \log \mathbb{E}_{s\sim \mathbb{P}_{\mathcal{M}(\mathcal{D}^\prime)}}\left\{\left(\frac{\operatorname{Pr}[\mathcal{M}(\mathcal{D})=s]}{\operatorname{Pr}[\mathcal{M}(\mathcal{D}^\prime)=s]}\right)^\alpha\right\}.
\end{aligned}
\end{equation}
\end{definition}

On the grounds of R\'enyi divergence, RDP is defined by the following definition.
\begin{definition}
\textup{(R\'enyi differential privacy \cite{mironov2017renyi}).} For $\alpha> 1$, $\varepsilon\geq0$, a randomized mechanism $\mathcal{M}:\mathcal{X}\mapsto \mathcal{Y}$ satisfies \emph{$(\alpha, \varepsilon)$-RDP} if, for every pair of adjacent datasets, $\mathcal{D}, \mathcal{D}^\prime \subseteq \mathcal{X}$, it holds that
\begin{equation}
\begin{aligned}
D_\alpha(\mathbb{P}_{\mathcal{M}(\mathcal{D})}||\mathbb{P}_{\mathcal{M}(\mathcal{D}^\prime)})\leq \varepsilon.
\end{aligned}
\end{equation}
\end{definition}

Note that RDP can be easily transformed into an equivalent characterization in terms of DP, as demonstrated by the following lemma.
\begin{lemma}
\label{lem:RDP2DP}
\textup{(From $(\alpha,\varepsilon)$-RDP to $(\epsilon,\delta)$-DP \cite{mironov2017renyi}).} If $\mathcal{M}$ is an $(\alpha,\varepsilon)$-RDP mechanism, it is also $(\varepsilon+\frac{\log1/\delta}{\alpha-1},\delta)$-DP for any $0<\delta<1$.
\end{lemma}

Based on the assumption that intermediate training models are not revealed, the technique \emph{privacy amplification by iteration} substantially improves the privacy guarantee analysis \cite{feldman2018privacy}, which is grounded on the concept of \emph{shifted R\'enyi divergence}, as follows.

\begin{definition}
\label{def:shifted}
    \textup{(Shifted R\'enyi divergence \cite{feldman2018privacy}).} Let $\boldsymbol{\mu}$, $\boldsymbol{\nu}$ be two random variables. Then, for any $z\geq0$ and $\alpha>1$, the $z$-shifted Rényi divergence is defined as
    \begin{equation}
    \mathcal{D}_\alpha^{(z)}(\mathbb{P}_{\boldsymbol{\mu}} || \mathbb{P}_{\boldsymbol{\nu}})=\inf_{\mathbb{P}_{\boldsymbol{\mu}^\prime}:W_{\infty}(\mathbb{P}_{\boldsymbol{\mu}},\mathbb{P}_{\boldsymbol{\mu}^\prime})\leq z}\mathcal{D}_\alpha(\mathbb{P}_{\boldsymbol{\mu}^\prime}||\mathbb{P}_{\boldsymbol{\nu}}),
    \end{equation}
    where $W_{\infty}(\cdot,\cdot)$ denotes the $\infty$-Wasserstein distance.\footnote{See Definition~\ref{def:wass} in Appendix~\ref{app:lemma}.}
\end{definition}

The privacy amplification by iteration utilizes the following lemma, which we frequently employ in the sequel.

\begin{lemma}
    \label{lemma:shifted-reduct}
    \textup{(Shift-reduction \cite{feldman2018privacy}).} Let $\boldsymbol{\mu}$, $\boldsymbol{\nu}$ be two random variables. Then, for any $a\geq0$ and $z\geq 0$, we have
    \begin{equation}
        \begin{aligned}
        \mathcal{D}_\alpha^{(z)}(\mathbb{P}_{\boldsymbol{\mu}} * \mathbb{P}_{\boldsymbol{\zeta}}  \|  \mathbb{P}_{\boldsymbol{\nu}} * \mathbb{P}_{\boldsymbol{\zeta}}) 
        \leq \mathcal{D}_\alpha^{(z+a)}(\mathbb{P}_{\boldsymbol{\mu}} \|  \mathbb{P}_{\boldsymbol{\nu}})+\frac{\alpha a^2}{2\sigma^2},
        \end{aligned}
    \end{equation}
where $\boldsymbol{\zeta}$ is a multi-variate Gaussian random variable with zero mean and covariance given by $\sigma^2 \boldsymbol{I}_d$, denoted by $\boldsymbol{\zeta}\sim\mathcal{N}(\boldsymbol{0},\sigma^2\boldsymbol{I}_d)$; and $\mathbb{P}_{\boldsymbol{\mu}} * \mathbb{P}_{\boldsymbol{\zeta}}$ denotes the distribution of the sum $\boldsymbol{\mu}+\boldsymbol{\zeta}$ with $\boldsymbol{\mu}$ and $\boldsymbol{\zeta}$ being drawn independently.
\end{lemma}




\subsection{DPSGD with Double Clipping (DPSGD-DC)}

\begin{algorithm}[ht]
   \caption{Differentially Private Stochastic Gradient Descent with Double Clipping (DPSGD-DC)}
   \label{alg:DPSGD}
\begin{algorithmic}
   \STATE {\bfseries Input:} Dataset $\mathcal{D}$, stochastic loss function $l_\xi(\boldsymbol{\theta}):\mathbb{R}^d\times \mathcal{D}\rightarrow \mathbb{R}$, learning rate $\eta$, noise scale $\sigma_{\text{DP}}$, dataset size $n$, batch size $b$, gradient norm bound $C$, parameter domain $\mathcal{K}$ with diameter $D$, number of iterations $T$;
   \STATE Initialize $\boldsymbol{\theta}_0\leftarrow \boldsymbol{0}$ and $t\leftarrow0$; \\
   \REPEAT
   \STATE \textbf{1) batch sampling:} \\
   take a random mini-batch $\mathcal{B}_t$ with sampling probability $q=b/n$;
   \STATE \textbf{2) compute and clip the gradients:} \\
   $\nabla \mathcal{L}_{\mathcal{B}_t}(\boldsymbol{\theta}_{t} ; \mathcal{D})\leftarrow\frac{1}{b} \sum_{\xi \in \mathcal{B}_t} \operatorname{clip}_C\left(\nabla l_{\xi}\left(\boldsymbol{\theta}_{t}\right)\right)$, \\
   where $\operatorname{clip}_C(\boldsymbol{x}) = \boldsymbol{x}\cdot\min(1,\frac{C}{\|\boldsymbol{x}\|})$; \\
   \STATE \textbf{3) update and project the parameters:} \\  $\boldsymbol{\theta}_{t+1}\leftarrow\Pi_{\mathcal{K}}(\boldsymbol{\theta}_{t}-\eta(\nabla \mathcal{L}_{B_t}\left(\boldsymbol{\theta}_{t};\mathcal{D}\right)+\boldsymbol{\zeta}_{t}))$,\\
   where $\Pi_{\mathcal{K}}(\boldsymbol{\theta})\!=\! \arg\min_{\boldsymbol{x}\in\mathcal{K}}\|\boldsymbol{\theta}-\boldsymbol{x}\|$ and $\boldsymbol{\zeta}_{t}\sim \!\mathcal{N}(\boldsymbol{0},\sigma_{\text{DP}}^2\boldsymbol{I}_d)$; \\
   \STATE \textbf{4) update the iteration counter:} \\
   $t \leftarrow t + 1$; \\
   \UNTIL{$t>T$}
   \STATE \textbf{\bfseries Output:} Final-round model parameters $\boldsymbol{\theta}_T$.
\end{algorithmic}
\end{algorithm}

In this paper, we also consider the DPSGD with both gradient clipping and parameter projection (Algorithm \ref{alg:DPSGD}), termed as \emph{DPSGD-DC}. This method begins with applying the vanilla SGD procedure using gradient clipping and Gaussian perturbation for model updates. Then, the updated parameters are projected into a bounded domain $\mathcal{K}=\{\boldsymbol{\theta}\in\mathbb{R}^d: \|\boldsymbol{\theta}\|\leq D\}$ as follows\footnote{A detailed discussion of the bounded domain assumption is provided in Appendix~\ref{discuss:bounded-domain}.} 
\begin{equation}
\label{eq:dpsgd}
\boldsymbol{\theta}_{t+1}=\Pi_\mathcal{K}\left(\boldsymbol{\theta}_{t}-\eta (\nabla \mathcal{L}_{B_t}\left(\boldsymbol{\theta}_{t};\mathcal{D}\right)+\boldsymbol{\zeta}_{t})\right),
\end{equation}
with 
\begin{equation}
    \Pi_{\mathcal{K}}(\boldsymbol{\theta})\!=\! \arg\min_{\boldsymbol{x}\in\mathcal{K}}\|\boldsymbol{\theta}-\boldsymbol{x}\|,
\end{equation}
where $\eta$ denotes the learning rate; $\boldsymbol{\zeta}_{t}\sim \mathcal{N}(\boldsymbol{0},\sigma_{\text{DP}}^2\boldsymbol{I}_d)$ denotes a multi-variate Gaussian random variable at iteration $t$; the loss function accrued on a model using data sample $\xi\in\mathcal{D}$ is defined as $l_\xi(\cdot):\mathcal{D}\times\mathbb{R}^d\mapsto\mathbb{R};$ and 
$\nabla \mathcal{L}_{B_t}\left(\boldsymbol{\theta}_{t} ; \mathcal{D}\right)$ is the clipped gradient of the loss function averaging over a mini-batch $\mathcal{B}_t$, i.e., $\nabla \mathcal{L}_{B_t}\left(\boldsymbol{\theta}_{t} ; \mathcal{D}\right)=\frac{1}{b} \sum_{\xi \in \mathcal{B}_t} \operatorname{clip}_C(\nabla l_\xi(\boldsymbol{\theta}_t))$, with $b=\left|\mathcal{B}_t\right|$ denoting the size of the mini-batch and $\operatorname{clip}_C(\boldsymbol{x}) = \boldsymbol{x}\cdot\min(1,\frac{C}{\|\boldsymbol{x}\|})$. Note that if $\mathcal{K}=\mathbb{R}^d$, it reduces to DPSGD with only gradient clipping, i.e., \emph{DPSGD-GC}.

\section{Main Theoretical Results}
\label{s3}


In this section, we construct RDP bounds to analyze privacy loss associated with releasing only the final-round model of DPSGD-GC and DPSGD-DC, respectively. To gain insights into how such privacy loss affects their respective training performance, we further provide utility analysis and derive the corresponding privacy-utility trade-offs.



\begin{assumption}
\label{A:smooth-stochastic}
\textup{($L$-smoothness of the loss function).} The loss function $l_\xi(\cdot): \mathcal{D}\times\mathbb{R}^d \mapsto \mathbb{R}$ is smooth with constant $L > 0$, if for any $\xi\in\mathcal{D}$ and all $\boldsymbol{\theta}, \boldsymbol{\theta}^{\prime} \in \mathbb{R}^d$, $l_\xi(\boldsymbol{\theta})$ is continuously differentiable, and its gradient $\nabla l_\xi(\cdot)$ in terms of $\boldsymbol{\theta}$ is $L$-Lipschitz, i.e.,  
\begin{equation}
    \left\|\nabla l_\xi(\boldsymbol{\theta})-\nabla l_\xi(\boldsymbol{\theta}^{\prime}\right)\| \leq L\left\|\boldsymbol{\theta}-\boldsymbol{\theta}^{\prime}\right\|.
\end{equation}
\end{assumption}

\subsection{Privacy Analysis of DPSGD}
In this subsection, first, we divide the original additive Gaussian noise $\boldsymbol{\zeta}_t\sim\mathcal{N}(\boldsymbol{0},\sigma_{\text{DP}}^2\boldsymbol{I}_d)$ (c.f. (\ref{eq:dpsgd})) into two parts: $\boldsymbol{\varrho}_t\sim\mathcal{N}(\boldsymbol{0},\beta\sigma_{\text{DP}}^2\boldsymbol{I}_d)$ and $\boldsymbol{\varsigma}_t\sim\mathcal{N}(\boldsymbol{0},(1-\beta)\sigma_{\text{DP}}^2\boldsymbol{I}_d)$. The first part, together with the clipped SGD update, constitutes the noisy update function, defined as
\begin{equation}
\label{eq:noise-update}
\psi(\boldsymbol{\theta}_t)\stackrel{\Delta}{=}\boldsymbol{\theta}_t-\eta\left(\frac{1}{b}\sum_{\xi \in \mathcal{B}_t} \operatorname{clip}_C\left(\nabla l_\xi\left(\boldsymbol{\theta}_t\right)\right)+\boldsymbol{\varrho}_t\right),    
\end{equation}
and the privacy loss of which is characterized by Lemma~\ref{lemma:smooth-reduct} presented shortly; the other part, $-\eta\boldsymbol{\varsigma}_t$, aims for reducing the shift amount of the shifted R\'enyi divergence leveraging Lemma~\ref{lemma:shifted-reduct}.


Next, we provide the following key lemma that provides the upper bound on the shifted R\'enyi divergence between the distributions of the noisy update function applied on two adjacent datasets with a smooth loss function.

\begin{lemma}
    \label{lemma:smooth-reduct}
    \textup{(Noisy smooth-reduction).} Let $\psi(\cdot)$ and $\psi^\prime(\cdot)$ be noisy update functions (c.f. (\ref{eq:noise-update})) of DPSGD based on adjacent datasets $\mathcal{D}$ and $\mathcal{D}^\prime$, respectively, and $n$ be the size of $\mathcal{D}$. If the loss function is $L$-smooth (Assumption~\ref{A:smooth-stochastic}), for any random variables $\boldsymbol{\mu}$ and $\boldsymbol{\nu}$, we have
    \begin{equation}
    \mathcal{D}_\alpha^{((1+\eta L)z)}(\mathbb{P}_{\psi(\boldsymbol{\mu})}||\mathbb{P}_{\psi^\prime(\boldsymbol{\nu})})
    \leq \ \mathcal{D}_\alpha^{(z)}(\mathbb{P}_{\boldsymbol{\mu}}||\mathbb{P}_{\boldsymbol{\nu}})+\frac{2\alpha C^2}{\beta nb\sigma_{\textup{DP}}^2}.
    \end{equation}
\end{lemma}
\begin{proofsketch}
    We summarize the proof steps as follows. First, we transform the shifted R\'enyi divergence to the standard R\'enyi divergence utilizing the smoothness of losses and equivalent definitions of $\infty$-Wasserstein distance. Next, the post-processing (Lemma~\ref{lem:post-process}) and partial convexity inequality (Lemma~\ref{lem:convex}) allow us to derive the privacy loss associated with SGD sampling. Finally, we apply the strong composition lemma (Lemma~\ref{lem:compos}) of RDP, and obtain the privacy loss associated with these two Gaussian distributions. The complete proof can be found in Appendix~\ref{proof:smooth-reduct}.
\end{proofsketch}

If we further assume the mini-batch size $b\leq \frac{n}{5}$, the RDP parameter $\alpha\leq\alpha^*(\frac{b}{n},\frac{b\sqrt{\beta}\sigma_{\textup{DP}}}{2C})$, and the per-dimension Gaussian noise scale $\sigma_{\textup{DP}}>\frac{8C}{b\sqrt{\beta}}$, a strengthened result can be obtained. Similar findings hold for all of our following theoretical results, and the complete version of Lemma~\ref{lemma:smooth-reduct} can be found in Appendix~\ref{proof:smooth-reduct}.

Note that this result generalizes the contraction-reduction lemma (see Lemma~\ref{lemma:contract-reduce}) \cite{feldman2018privacy} and its variants \cite{altschuler2022privacy,altschuler2024privacy} in existing literature, which all rely on the convexity of the loss function to ensure that the update function is contractive.\footnote{ A function is said to be contractive if it is $1$-Lipschitz.} It characterizes the privacy dynamics of shifted R\'enyi divergence for noisy stochastic updates with general smooth loss functions. Based on this building block, we are ready to present the privacy guarantees for DPSGD-GC and DPSGD-DC, respectively.

\begin{theorem}
\label{thm:privacy-GC}
\textup{(Privacy guarantee for DPSGD-GC).} Given the number of total  iterations $T$, dataset size $n$, batch size $b$, stepsize $\eta$, $\alpha> 1$, gradient-clipping threshold $C$, and noise scale $\sigma_{\textup{DP}}$, if the loss function is $L$-smooth (Assumption~\ref{A:smooth-stochastic}), then the DPSGD-GC algorithm satisfies $(\alpha,\varepsilon)$-RDP for
\begin{equation}
    \varepsilon = \mathcal{O}\left(\frac{\alpha C^2}{nb\sigma_{\textup{DP}}^2}T\right).
\end{equation}
\end{theorem}

\begin{proofsketch}
We establish our result utilizing the recursive hypothesis from $T$ to $0$ and the flexibility of the shifted R\'enyi divergence. For the base case at $t=0$, we have $\mathcal{D}_\alpha^{(z_0)}(\mathbb{P}_{\boldsymbol{\theta}_0}||\mathbb{P}_{\boldsymbol{\theta}_0^\prime})=0$, since the initialization satisfy $\boldsymbol{\theta}_0=\boldsymbol{\theta}^\prime_0$. For the recursive step, we apply Lemma~\ref{lemma:smooth-reduct} and subsequently reduce Lemma~\ref{lemma:shifted-reduct} with auxiliary variables to derive the recurrence relationship. Finally, by tracking the privacy loss across all iterations, we derive an upper bound on the RDP loss. The complete proof can be found in Appendix~\ref{proof:privacy-GC}.
\end{proofsketch}

\begin{theorem}
\label{thm:privacy-DC}
\textup{(Privacy guarantee for DPSGD-DC).} Given the number of total iterations $T$, dataset size $n$, batch size $b$, stepsize $\eta$, $\alpha>1$, gradient-clipping threshold $C$, diameter of the model-parameter domain $D$, and noise scale $\sigma_{\textup{DP}}$, if the loss function is $L$-smooth (Assumption~\ref{A:smooth-stochastic}), then the DPSGD-DC algorithm satisfies $(\alpha,\varepsilon)$-RDP for
\begin{equation}
    \varepsilon=\mathcal{O}\left(\frac{\alpha C^2}{nb\sigma_{\textup{DP}}^2}\min\left\{T, \frac{(1+\eta L)^2nbD^2}{\eta^2C^2}\right\}\right).
\end{equation}
\end{theorem}

\begin{proofsketch}
    This proof follows a similar approach to Theorem~\ref{thm:privacy-GC} but terminates the recursion early at iteration $\tau$. By judiciously setting the shift amount $z_\tau=D$ at $t=\tau$, we obtain $\mathcal{D}_\alpha^{(z_\tau)}(\mathbb{P}_{\boldsymbol{\theta}_\tau}||\mathbb{P}_{\boldsymbol{\theta}_\tau^\prime})=0$ due to  the bounded domain assumption. Finally, by appropriately setting the values of the auxiliary shift variables, we derive the converged privacy loss, as detailed in Appendix~\ref{proof:privacy-DC}.
\end{proofsketch}

\begin{remark}
    Comparing Theorem~\ref{thm:privacy-DC} to \ref{thm:privacy-GC}, we observe that the privacy loss for DGSGD-DC can still converge to a constant for non-convex smooth loss function when the model-parameter domain is bounded by $D$. However, unlike the convex cases \cite{altschuler2022privacy, altschuler2024privacy}, where the upper bound scales linearly with $D$, the non-convex setting yields a bound that scales quadratically with $D$. This suggests that upper bounds on RDP for bounded-domain scenarios are inherently looser without the assumption on convexity, which is consistent with intuition and recent findings \cite{kong2024privacy, chien2024convergent}.
\end{remark}

\subsection{Utility Analysis of DPSGD}
In this subsection, we provide upper bounds on utility functions that reflect convergence behavior of DPSGD-GC and DPSGD-DC under $(\alpha,\varepsilon)$-RDP constraint, respectively. All bounds are expressed as expectations over the randomness of SGD sampling and Gaussian noise. 

\begin{assumption}
\label{A:bound-SGD}
\textup{(Bounded SGD variance).} The gradient $\nabla l_{\xi}(\cdot)$ has bounded variance. That is, for all $\boldsymbol{\theta}\in\mathbb{R}^d$, we have
\begin{equation}
    \mathbb{E}_{\xi\sim\mathbb{P}_{\mathcal{D}}}\left[\|\nabla l_\xi(\boldsymbol{\theta})-\nabla l(\boldsymbol{\theta})\|^2\right]\leq \sigma_{\text{SGD}}^2.
\end{equation}
\end{assumption}

Note that Assumption~\ref{A:smooth-stochastic} implies that the population risk function $l(\cdot)=\mathbb{E}_{ \xi}[l_{\xi}(\cdot) ]$ is also smooth with constant $L>0$. Based on Assumption~\ref{A:smooth-stochastic} and \ref{A:bound-SGD}, the following lemma provides an upper bound on the minimum expected norm of the gradient for DPSGD-GC with smooth population risk functions.
\begin{lemma}
\label{lem:converge-GC}
\textup{(Utility bound for DPSGD-GC \cite{koloskova2023revisiting}).} Assume that the population risk function $l(\cdot)$ has smoothness parameter $L$ and SGD variance bounded from above by $\sigma_{\textup{SGD}}^2$ (Assumption~\ref{A:bound-SGD}). The DPSGD-GC running for $T$ iterations with step-size $\eta\leq \frac{1}{9L}$ has the minimum expected norm of the population risk's gradient upper bounded by
\begin{equation}
\begin{aligned}
&\min_{t\in[0,T]} \mathbb{E}\left[\|\nabla l(\boldsymbol{\theta}_t)\|\right] \\
&\leq \mathcal{O} \biggl(\frac{1}{\eta CT}+\frac{1}{\sqrt{\eta T}}+\min \bigl(\sigma_{\textup{SGD}}, \frac{\sigma_{\textup{SGD}}^2}{C}\bigr)\\
&+\sqrt{\eta L} \frac{\sigma_{\textup{SGD}}}{\sqrt{b}}+\frac{dL \eta}{C} \sigma_{\textup{DP}}^2+\sqrt{d L \eta} \sigma_{\textup{DP}}\biggr),
\end{aligned}
\end{equation}
\end{lemma}

\begin{proposition}
\label{coro:utility-GC}
\textup{(Privacy-utility trade-off for DPSGD-GC).} Assuming that the conditions in Lemma~\ref{lem:converge-GC} are satisfied, for DPSGD-GC with $L$-smooth population risk and $\sigma_{\textup{DP}}^2=\mathcal{O}\bigl(\frac{\alpha C^2 T}{\varepsilon nb}\bigr)$, we have the following results:
\begin{equation}
    \begin{aligned}
        &\min_{t\in[0,T]} \mathbb{E}\left[\|\nabla l(\boldsymbol{\theta}_t)\|\right] \\
        &\leq \mathcal{O} \biggl(\frac{1}{\eta CT}+\frac{1}{\sqrt{\eta T}}+\min \left\{\sigma_{\textup{SGD}}, \frac{\sigma_{\textup{SGD}}^2}{C}\right\}\\
        &+\sqrt{\eta L} \frac{\sigma_{\textup{SGD}}}{\sqrt{b}}+\frac{\alpha dL\eta CT}{\varepsilon n b}+\frac{\sqrt{\alpha dL\eta T}C}{\sqrt{\varepsilon nb}}\biggr).
    \end{aligned}
    \end{equation}
   
\end{proposition}

Following this proposition, we can derive the achievable privacy-utility trade-off for DPSGD-GC as $\mathcal{O} (\max\{\frac{dL\log(1/\delta)}{\epsilon^2n^2},\sigma_{\text{SGD}}^{4/3}[\frac{dL\log(1/\delta)}{\epsilon^2n^2}]^{1/3}\})$, and we defer the details to 
Appendix~\ref{discuss:privacy-utility-GC}.

\begin{assumption}
\label{A:strong-convex}
\textup{($\mu$-strongly convex).} The population risk function $l(\cdot)$ is strongly convex with constant $\mu > 0$ if and only if the following inequality holds for all $\boldsymbol{\theta}, \boldsymbol{\theta}^{\prime} \in \mathbb{R}^d$, i.e.,
\begin{equation}
    \left[\nabla l(\boldsymbol{\theta})-\nabla l\left(\boldsymbol{\theta}^{\prime}\right)\right]^{T}\left(\boldsymbol{\theta}-\boldsymbol{\theta}^{\prime}\right) \geq \mu\left\|\boldsymbol{\theta}-\boldsymbol{\theta}^{\prime}\right\|^2. 
\end{equation}
\end{assumption}

For DPSGD-DC, in the special case of smooth and strongly convex population risk, we obtain the following upper bound that describes the minimum expected square root of the optimality gap.
\begin{theorem}
\label{thm:converge-DC}
\textup{(Utility bound for DPSGD-DC).} Assume that the population risk function $l(\cdot)$ has smoothness parameter $L$, strongly convex parameter $\mu$ (Assumption~\ref{A:strong-convex}), SGD variance bounded from above by $\sigma_{\textup{SGD}}^2$ (Assumption~\ref{A:bound-SGD}), and $\boldsymbol{\theta}^*=\arg\min_{\boldsymbol{\theta}}l(\boldsymbol{\theta})\in \operatorname{int} \mathcal{K}$. The DPSGD-DC running for $T$ iterations with step-size $\eta\leq \frac{9}{20L}$ has a minimum expected square root of the optimality gap upper bounded by
\begin{equation}
\begin{aligned}
    &\min_{t\in[0,T]} \mathbb{E}\left[\sqrt{l(\boldsymbol{\theta}_t)-l(\boldsymbol{\theta}^*)}\right]\\ &\leq \mathcal{O}\biggl(\frac{\sqrt{L}D^2}{\eta C T}+\frac{D}{\sqrt{\eta  T}}+ \min\left\{\frac{L^{3/4}}{\mu^{5/4}}\sigma_{\textup{SGD}},\sqrt{\frac{\sigma_{\textup{SGD}}^3}{\mu C}}\right\} \\
    &+\frac{\sqrt{\eta}\sigma_{\textup{SGD}}}{\sqrt{b}}+\frac{d\eta\sigma_{\textup{DP}}^2\sqrt{L}}{C}+\sqrt{d\eta}\sigma_{\textup{DP}}\biggr).
\end{aligned}
\end{equation}
\end{theorem}
\begin{proofsketch}
    We present an intuitive sketch below, with the complete proof provided in Appendix~\ref{proof:converge-DC}. The primary challenges stem from the gradient clipping operation, the SGD procedure, and the parameter projection step. To address these, we divide the proof into several cases. For instance, in the case where the clipping threshold $C\leq10\sigma_{\text{SGD}} (\frac{L}{\mu})^{\frac{1}{2}}$ and $\|\nabla l(\boldsymbol{\theta}_t)\|\geq 35\sigma_{\text{SGD}} (\frac{L}{\mu})^{\frac{3}{4}}$, first, we leverage the non-expansiveness of the projection operator to analyze the impact of the parameter projection. Next, we introduce an auxiliary clipping factor $\gamma_\xi=\min (1, \frac{C}{\|\nabla l_\xi(\boldsymbol{\theta}_t)\|})$ and apply the Markov inequality to quantify the effects of clipped SGD. By careful step-size design and manipulations, we obtain a recurrence relation that characterizes the evolution over two successive time steps. Finally, by averaging over $t$, we derive an upper bound for this case. The proof for other cases follows a similar procedure but incorporates different auxiliary variables tailored to the true gradient. By summing up all cases, we derive an upper bound on the error for DPSGD-DC.
\end{proofsketch}

\begin{remark}
    Note that our results adopt a different utility metric other than the more commonly used $\frac{1}{T}\sum_{t=1}^T \mathbb{E}[l(\boldsymbol{\theta}_t)-l(\boldsymbol{\theta}^*)]$, as this metric better facilitates our analysis and gaining insights into the results.
\end{remark}

Using the RDP guarantees in Theorem~\ref{thm:privacy-DC}, we immediately obtain the following results.

\begin{proposition}
\label{coro:utility-DC}
\textup{(Privacy-utility trade-off for DPSGD-DC).} Assuming that the conditions in Theorem~\ref{thm:converge-DC} are satisfied, for DPSGD-DC with $L$-smooth and $\mu$-strongly convex population risk, $\sigma_{\textup{DP}}^2=\mathcal{O}\bigl(\frac{\alpha C^2}{\varepsilon nb}\min\{T, \overline{T}\}\bigr)$ and $\overline{T}=\frac{(1+\eta L)^2nbD^2}{\eta^2C^2}$, we have the following results:
%
\begin{equation}
\begin{aligned}
    &\min_{t\in[0,T]} \mathbb{E}\left[\sqrt{l(\boldsymbol{\theta}_t)-l(\boldsymbol{\theta}^*)}\right]\leq \mathcal{O}\biggl(\frac{\sqrt{L}D^2}{\eta C T} +\frac{D}{\sqrt{\eta T}}\\
    &+ \min\left\{\frac{L^{3/4}}{\mu^{5/4}}\sigma_{\textup{SGD}},\sqrt{\frac{\sigma_{\textup{SGD}}^3}{\mu C}}\right\}+\frac{\sqrt{\eta}\sigma_{\textup{SGD}}}{\sqrt{b}} \\
    &+\frac{\alpha d\eta C\sqrt{L}}{\varepsilon nb} \min\bigl\{T, \overline{T}\bigr\} +\sqrt{\frac{\alpha d\eta}{\varepsilon nb} \min\bigl\{T, \overline{T}\bigr\} } C\biggr).
\end{aligned}
\end{equation}


\end{proposition}


Proposition~\ref{coro:utility-DC} suggests that the utility bound for DPSGD-DC comprises six terms. The first two terms capture optimization-related factors, reflecting the influence of clipping and projection in convergence behavior. The third term accounts for the inherent bias introduced by gradient clipping, while the fourth term reflects the effect due to SGD sampling. Finally, the last two terms quantify the impact of the injected DP noise. 

Following Proposition~\ref{coro:utility-DC}, we can derive the achievable privacy-utility trade-off for DPSGD-DC as 
\begin{equation}
\begin{aligned}
\mathcal{O} \biggl(\max\biggl\{\frac{D^2dL\log(1/\delta)}{\epsilon^2n^2}&,\frac{\sigma_{\text{SGD}}^{3/2}D^{1/2}}{\mu^{1/2}}[\frac{dL\log(1/\delta)}{\epsilon^2n^2}]^{1/4}\\
&,\frac{\sigma_{\text{SGD}}D \sqrt{d\log(1/\delta)}}{\sqrt{b}\epsilon}\biggr\}\biggr),    
\end{aligned}
\end{equation}
and we provide the details in
Appendix~\ref{discuss:privacy-utility-DC}.


\begin{remark}
    Unlike previous works on DPSGD-DC \cite{altschuler2022privacy,altschuler2024privacy, kong2024privacy, chien2024convergent}, which mainly focused on RDP analysis, our results explicitly demonstrate how gradient clipping and projection affect the utility of DPSGD-DC. As seen from Theorem~\ref{thm:privacy-DC}, we establish an eventually constant upper bound on privacy loss dependent on the bounded domain diameter $D$, which yields tighter privacy guarantees with smaller $D$. Meanwhile, Proposition~\ref{coro:utility-DC} demonstrates that when the conditions in Theorem~\ref{thm:converge-DC} are satisfied, a smaller $D$ also leads to a lower upper bound on the utility, thus enhancing the privacy-utility trade-off.\footnote{A brief discussion on selecting $D$ is provided in Appendix~\ref{discuss:select-D}.}
\end{remark}

\begin{figure}[!t]
  \centering
  \subfigure[]{\includegraphics[width=0.44\textwidth]{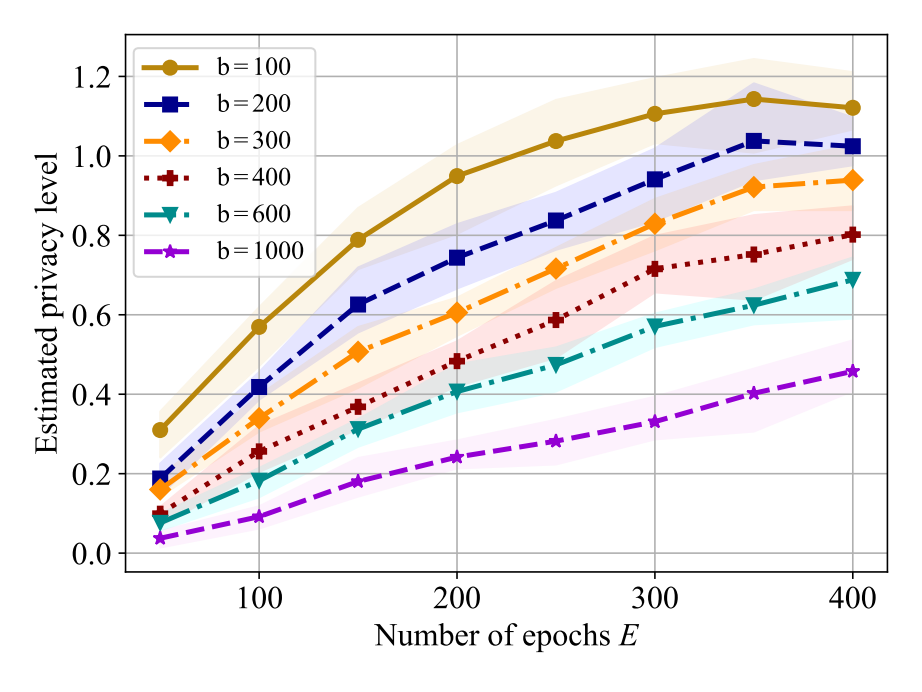}}
  \subfigure[]{\includegraphics[width=0.47\textwidth]{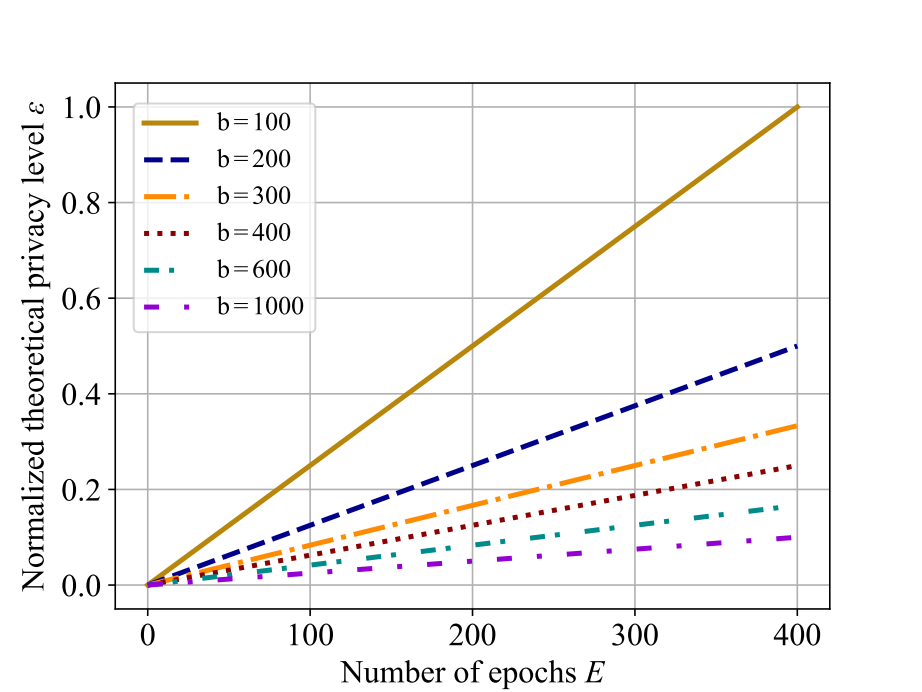}}
  \caption{The evolution of the: (a) estimated and (b) normalized theoretical privacy level during DPSGD-GC with different batch sizes. The shaded error bars correspond to intervals covering $95$\% of the realized values, obtained from the $10$ Monte Carlo trials. Note that the privacy bounds in terms of the number of epochs, $E$, can be derived by substituting $T=\lceil \frac{n}{b}\rceil E$ into our main results.}
\label{fig:batch}
\end{figure}

\begin{figure}[!t]
\centering
\subfigure[]{\includegraphics[width=0.46\textwidth]{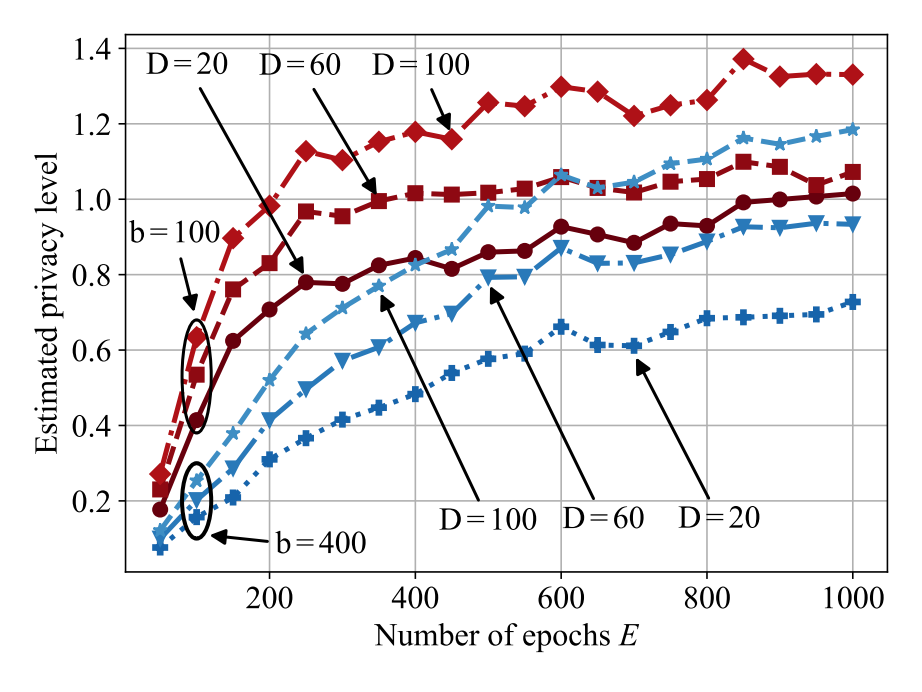}}
\subfigure[]{\includegraphics[width=0.49\textwidth]{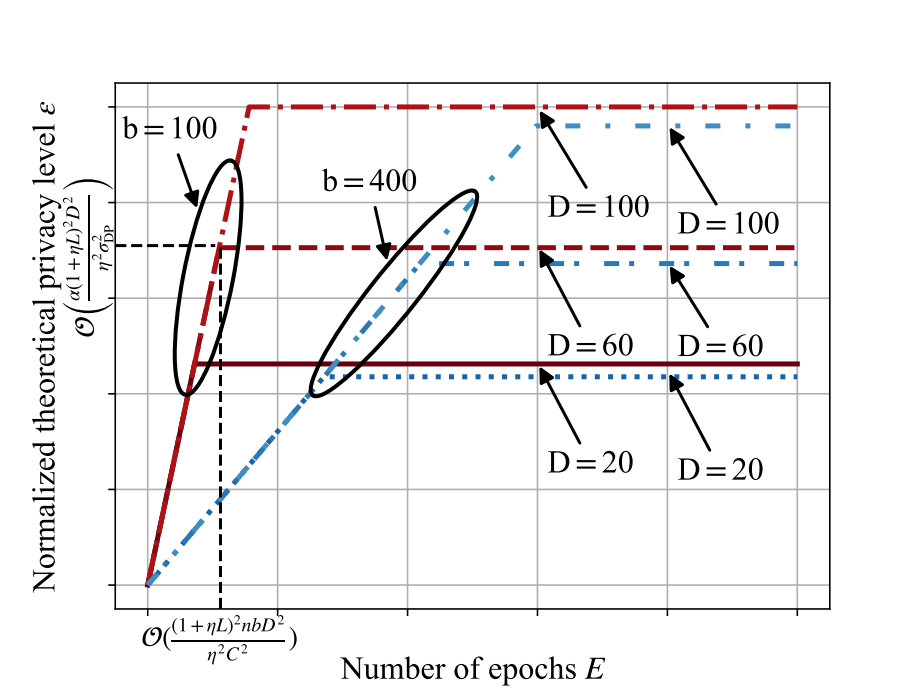}}
\caption{The evolution of the privacy level during DPSGD-DC with different diameters of the bounded domain: (a) estimated by MIA and (b) normalized theoretical privacy level $\varepsilon$ with $\alpha=1.1$. The red and blue lines correspond to the cases with batch sizes of $100$ and $400$, respectively.}
\label{fig:D}
\end{figure}

\section{Experiment Evaluation}
\label{s4}
In this section, we present empirical results estimating the privacy level via MIA. Following \citet{kairouz2015composition}, we estimate $(\epsilon,\delta)$-DP using the false positive rate (FPR) and false negative rate (FNR) of its attack model on the test data, applying the following formula: 
\begin{equation}
    \hat{\epsilon} = \max \biggl\{\log \frac{1-\delta-\text{FPR}}{\text{FNR}},\log \frac{1-\delta-\text{FNR}}{\text{FPR}}\biggr\}.
\end{equation}


We emphasize that the MIA serves primarily as a tool to provide a lower bound on privacy, capturing the trend with privacy level changes and validating the consistency of the theoretical bounds. By comparing experimental results with the theoretical trends, we aim for demonstrating the reasonableness of the derived privacy bounds, rather than offering an exact measure of privacy leakage. This approach allows us to examine how privacy evolves with varying experimental conditions under the same privacy attack. Additional information on the MIA setting and implementation details can be found in Appendix~\ref{experiment-details}.


\paragraph{Effect of the Batch Size.}
We evaluate DPSGD-GC with various batch sizes $b\in \{100, 200, 300, 400, 600, 1000\}$. Figure~\ref{fig:batch} illustrates the evolution of the estimated and theoretical privacy level. We also report the evolution of training loss in Figure~\ref{fig:loss-GC} in Appendix~\ref{app:miss-fig}. As expected, the estimated privacy level 
$\hat{\epsilon}$ increases with the number of epochs, and larger batch sizes provide stronger privacy protection. These observations align with our theoretical results in Theorem~\ref{thm:privacy-GC}. Additionally, DPSGD-GC converges more slowly with larger batch sizes, consistent with Lemma~\ref{lem:converge-GC}, further highlighting the trade-off between privacy and utility.


\paragraph{Effect of the Bounded Domain Diameter.}
We then conduct experiments on DPSGD-DC using various diameters for the bounded domain $D\in\{20,60,100\}$ with batch sizes of $100$ and $400$, respectively. The estimated and theoretical privacy parameter of DPSGD-DC for different bounded domain diameters $D$ is shown in Figure~\ref{fig:D}. As can be observed, DPSGD-DC provides stronger privacy guarantees with a smaller $D$, as limiting the domain diameter restricts the range of parameter variations to a narrower interval. Additionally, privacy leakage tends to stabilize as the number of epochs increases. This observation is not surprising, as the bounded domain assumption provides a constant upper privacy bound for DPSGD-DC, as shown in Theorem~\ref{thm:privacy-DC}.

\section{Conclusions}
\label{s5}

In this paper, we have rigorously analyzed the privacy and utility guarantee of DPSGD, considering both gradient clipping (DPSGD-GC) and double clipping (DPSGD-DC). Our analysis extends the existing privacy bounds of DPSGD to general smooth and non-convex problems without relying on other assumptions. While previous works have focused solely on privacy characterization, we have also derived utility bounds corresponding to our RDP guarantees. This dual characterization admits a more comprehensive understanding of the privacy-utility trade-offs in DPSGD, providing valuable insights for developing more effective differentially private optimization algorithms.


\appendix

\section{Preliminaries}
\label{app:lemma}

\subsection{Privacy Amplification by Iteration}

The privacy amplification by iteration framework relies on two key lemmas, focusing specifically on Gaussian noise, which suffices for our purposes. First, Lemma~\ref{lemma:shifted-reduct} stated in the main body will prove to be useful for analyzing shifted R\'enyi divergence between two distributions convolved with Gaussian noise.

The contraction-reduction lemma provides an upper bound on the shifted R\'enyi divergence between the distributions of two pushforwards through similar contraction maps.

\begin{lemma}[Contraction-reduction \cite{feldman2018privacy}]
    \label{lemma:contract-reduce}
    Let $\boldsymbol{\mu}$, $\boldsymbol{\nu}$ be two random variables. Suppose $\phi(\cdot)$, $\phi^\prime(\cdot)$ are two contractive functions and let $\sup_{\boldsymbol{x}} \|\phi(\boldsymbol{x})-\phi^\prime(\boldsymbol{x})\| \leq s$. Then, for any $z\geq0$, we have
    \begin{equation}
        \mathcal{D}_\alpha^{(z+s)}(\mathbb{P}_{\phi(\boldsymbol{\mu})}||\mathbb{P}_{\phi^\prime(\boldsymbol{\nu})})\leq\mathcal{D}_\alpha^{(z)}(\mathbb{P}_{\boldsymbol{\mu}}||\mathbb{P}_{\boldsymbol{\nu}}).
    \end{equation}
\end{lemma}

Altschuler et al. \cite{altschuler2022privacy, altschuler2024privacy} provide a mild generalization of this lemma, extending it to be suitable for random contraction maps.

\subsection{Standard Facts}

To prove our main results, we first introduce the following standard facts.

\begin{lemma}[Markov inequality]
\label{lem:markov}
If $x$ is a non-negative random variable and $a > 0$, then we have
\begin{equation}
    \operatorname{Pr}(x\geq a)\leq \frac{\mathbb{E}(x)}{a}.
\end{equation}
\end{lemma}

\begin{lemma}[Projection operator is non-expansive]
\label{lem:project}
For any $\boldsymbol{x}, \boldsymbol{y}\in \mathbb{R}^d$, we have
\begin{equation}
    \|\Pi_\mathcal{K}(\boldsymbol{x})-\Pi_\mathcal{K}(\boldsymbol{y})\| \leq \|\boldsymbol{x}-\boldsymbol{y}\|,
\end{equation}
where $\Pi_{\mathcal{K}}(\boldsymbol{x})= \arg\min_{\boldsymbol{z}\in\mathcal{K}}\|\boldsymbol{x}-\boldsymbol{z}\|$ denotes the projection operator.
\end{lemma}

\begin{lemma}
\label{lem:(a+b)2}
    For any $a,b \in \mathbb{R}$, it holds that
    \begin{equation}
        (a+b)^2\leq 2a^2+2b^2.
    \end{equation}
\end{lemma}

\begin{lemma}
\label{lem:x2x}
    For any $a,b \geq0$, it holds that
    \begin{equation}
        a\leq \frac{a^2}{2b}+\frac{b}{2}.
    \end{equation}
\end{lemma}

\begin{lemma}
\label{lem:sqrt(a+b)}
    For any $a,b \geq0$, it holds that
    \begin{equation}
        \sqrt{a+b}\leq \sqrt{a}+\sqrt{b}.
    \end{equation}
\end{lemma}

\subsection{Lemmas for Privacy Analysis}

We then provide some supporting lemmas that are useful for our proof. First, the following lemma provides equivalent characterizations of the $\infty$-Wasserstein distance:
\begin{definition}[$\infty$-Wasserstein distance \cite{villani2009wasserstein}]
\label{def:wass}
    The $\infty$-Wasserstein distance between two distributions $\mathbb{P}_{\boldsymbol{\mu}}$ and $\mathbb{P}_{\boldsymbol{\nu}}$ is defined as
\begin{equation}
    W_{\infty}(\mathbb{P}_{\boldsymbol{\mu}}, \mathbb{P}_{\boldsymbol{\nu}})=\inf_{\gamma \in \Gamma(\mathbb{P}_{\boldsymbol{\mu}}, \mathbb{P}_{\boldsymbol{\nu}})} \operatorname{ess \ sup}_{(\boldsymbol{x}, \boldsymbol{y}) \sim \gamma}\|\boldsymbol{x}-\boldsymbol{y}\|,
\end{equation}
where $(\boldsymbol{x},\boldsymbol{y})\sim \gamma$ indicates that $(\boldsymbol{x},\boldsymbol{y})$ follows the joint distribution $\gamma$, and $\Gamma(\mathbb{P}_{\boldsymbol{\mu}},\mathbb{P}_{\boldsymbol{\nu}})$ represents the collection of all joint distributions with $\mathbb{P}_{\boldsymbol{\mu}}$ and $\mathbb{P}_{\boldsymbol{\nu}}$ being as marginals of each random variable.
\end{definition} 

\begin{lemma}[Equivalent definitions of $\infty$-Wasserstein distance \cite{feldman2018privacy}]
\label{lem:W-distance} 
The following are equivalent for any distributions $\mathbb{P}_{\boldsymbol{\mu}}$ and $\mathbb{P}_{\boldsymbol{\nu}}$:
\begin{enumerate}
\item $W_\infty(\mathbb{P}_{\boldsymbol{\mu}},\mathbb{P}_{\boldsymbol{\nu}})\leq s$.
\item There exist jointly distributed random variables $(\boldsymbol{u}, \boldsymbol{v})$ such that $\boldsymbol{u}\sim \mathbb{P}_{\boldsymbol{\mu}}$, $\boldsymbol{v}\sim \mathbb{P}_{\boldsymbol{\nu}}$, and $\operatorname{Pr}[\|\boldsymbol{u}-\boldsymbol{v}\|\leq s]=1$.
\item There exist jointly distributed random variables $(\boldsymbol{u},\boldsymbol{w})$ such that $\boldsymbol{u}\sim \mathbb{P}_{\boldsymbol{\mu}}$, $\boldsymbol{u}+\boldsymbol{w}\sim \mathbb{P}_{\boldsymbol{\nu}}$, and $\operatorname{Pr}[\|\boldsymbol{w}\|\leq s]=1$.
\end{enumerate}
\end{lemma}

Further, R\'enyi divergence has several fundamental properties, including non-negative and non-decreasing with respect to $\alpha$. One such key property is the post-processing inequality, which we formalize in the following lemma:
\begin{lemma}[Post-processing inequality \cite{van2014renyi}]
\label{lem:post-process} 
Let $\boldsymbol{\mu}$ and $\boldsymbol{\nu}$ be two random variables. Then, for any (possibly random) function $\psi$ and $\alpha\geq0$, we have
\begin{equation}
\mathcal{D}_\alpha(\mathbb{P}_{\psi(\boldsymbol{\mu})}||\mathbb{P}_{\psi(\boldsymbol{\nu})})\leq\mathcal{D}_\alpha(\mathbb{P}_{\boldsymbol{\mu}}||\mathbb{P}_{\boldsymbol{\nu}}).
\end{equation}
\end{lemma}
The following lemma demonstrates the partial convexity property of R\'enyi divergence.
\begin{lemma}[Partial convexity in the second argument \cite{van2014renyi}]
\label{lem:convex}
    Let $\boldsymbol{\mu}$, $\boldsymbol{\nu}_1$, and $\boldsymbol{\nu}_2$ be three random variables. Then, for any $\alpha\geq0$, the R\'enyi divergence is convex in its second argument, that is, the following inequality holds
    \begin{equation}
    \begin{aligned}
    &\mathcal{D}_\alpha(\mathbb{P}_{\boldsymbol{\mu}}||(1-\lambda)\mathbb{P}_{\boldsymbol{\nu}_1}+\lambda\mathbb{P}_{\boldsymbol{\nu}_2})\\
    &\leq (1-\lambda)\mathcal{D}_\alpha(\mathbb{P}_{\boldsymbol{\mu}}||\mathbb{P}_{\boldsymbol{\nu}_1})+\lambda\mathcal{D}_\alpha(\mathbb{P}_{\boldsymbol{\mu}}||\mathbb{P}_{\boldsymbol{\nu}_2}),
    \end{aligned}
    \end{equation}
    for any $0\leq\lambda\leq 1$.
\end{lemma}
The third property is the composition of two RDP mechanisms, as follows:
\begin{lemma}[Strong composition for R\'enyi divergence \cite{mironov2017renyi}]
\label{lem:compos}
    Let $\mathcal{M}=\left[\mathcal{M}_1(\mathcal{D}),\mathcal{M}_2\left(\mathcal{D}\right)\right]$ be a sequence of two randomized mechanisms. Then, for any $\alpha>1$, we have
    \begin{equation}
    \begin{aligned}
    &\mathcal{D}_\alpha(\mathbb{P}_{\mathcal{M}(\mathcal{D})}||\mathbb{P}_{\mathcal{M}(\mathcal{D}^\prime)})\\
    &\leq \sup_{\boldsymbol{v}} \mathcal{D}_\alpha(\mathbb{P}_{\mathcal{M}_2(\mathcal{D})|\mathcal{M}_1(\mathcal{D})=\boldsymbol{v}}||\mathbb{P}_{\mathcal{M}_2(\mathcal{D}^\prime)|\mathcal{M}_1(\mathcal{D}^\prime)=\boldsymbol{v}})\\
    &+\mathcal{D}_\alpha(\mathbb{P}_{\mathcal{M}_1(\mathcal{D})}||\mathbb{P}_{\mathcal{M}_1(\mathcal{D}^\prime)}),
    \end{aligned}
    \end{equation}
    where $\mathcal{D}$ and $\mathcal{D}^\prime$ are two adjacent datasets.
\end{lemma}

Finally, we briefly introduce a concise statement of the RDP analysis for the sampled Gaussian mechanism (SGM), which can simplify our analysis under additional parameter assumptions. The SGM is a composition of subsampling and the additive Gaussian noise, whose formal definition is given as follows:
\begin{definition}[Sampled Gaussian mechanism]
Let $f(\cdot):\mathcal{D}\rightarrow\mathbb{R}^d$ and $\mathcal{S}$ be a sample from $[n]$ where each $i\in [n]$ is chosen independently with probability $0<q\leq1$ and $n=|\mathcal{D}|$. The SGM, parameterized with the noise scale $\sigma>0$, is defined as 
\begin{equation}
\begin{aligned}
\mathcal{M}_{\text{SGM}}(\mathcal{D}) = \sum_{i\in \mathcal{S}} f(\mathcal{D}_i) + \mathcal{N}(\boldsymbol{0},\sigma^2\boldsymbol{I}_d),
\end{aligned}
\end{equation}
where $\mathcal{D}_i$ denotes a single element of the dataset $\mathcal{D}$. For R\'enyi parameter $\alpha>1$, the R\'enyi divergence of SGM is defined as
\begin{equation}
\begin{aligned}
    &\mathcal{D}_\alpha^{\text{SGM}}(q, \sigma)\\
    &=\mathcal{D}_\alpha\left(\mathcal{N}(0,\sigma^2)||(1-q)\mathcal{N}(0,\sigma^2)+q\mathcal{N}(1,\sigma^2)\right).
\end{aligned}
\end{equation}
\end{definition}

The following lemma provides a closed-form upper bound on $\mathcal{D}_\alpha^{\text{SGM}}(q, \sigma)$:
\begin{lemma}[Upper bound on R\'enyi divergence of the SGM \cite{mironov2019r}]
\label{lem:SGM}
    If $q\leq 1/5$, $\sigma>4$, and $\alpha\leq\alpha^*(q,\sigma)$, i.e.,
    \begin{equation}
    \begin{aligned}
        \alpha &\leq K \sigma^2/2-2\log \sigma, \\
        \alpha &\leq \frac{\left(K^2 \sigma^2 / 2-\log 5-2\log \sigma\right)}{\left(K+\log (q \alpha)+1 /(2 \sigma^2)\right)},
    \end{aligned}
    \end{equation}
    where $K=\log(1+1/(q(\alpha-1)))$, then $\mathcal{D}_\alpha^{\text{SGM}}(q, \sigma)\leq \frac{2\alpha q^2}{\sigma^2}$.
\end{lemma}

\subsection{Lemmas for Utility Analysis}

The following results provide several equivalent characterizations of the $L$-smoothness property for functions that are also convex:
\begin{lemma}[Implications of smoothness \cite{beck2017first}]
\label{lem:smooth}
Let $f(\cdot):\mathbb{R}^d\rightarrow\mathbb{R}$ be a convex function, differentiable over $\mathbb{R}^d$. Then, the following claims are equivalent:
\begin{itemize}
    \item $f(\cdot)$ is $L$-smooth.
    \item $f(\boldsymbol{y})\leq f(\boldsymbol{x})+\nabla f(\boldsymbol{x})^T(\boldsymbol{y}-\boldsymbol{x})+\frac{L}{2}\|\boldsymbol{x}-\boldsymbol{y}\|^2$ for all $\boldsymbol{x}, \boldsymbol{y}\in\mathbb{R}^d$.
    \item $f(\boldsymbol{y})\geq f(\boldsymbol{x})+\nabla f(\boldsymbol{x})^T(\boldsymbol{y}-\boldsymbol{x})+\frac{1}{2L}\|\nabla f(\boldsymbol{x})-\nabla f(\boldsymbol{y})\|^2$ for all $\boldsymbol{x}, \boldsymbol{y}\in\mathbb{R}^d$.
\end{itemize}
\end{lemma}

Lemma~\ref{lem:strong-convex} below describes some useful properties of strong convexity.
\begin{lemma}[Implications of strong convexity \cite{beck2017first}]
\label{lem:strong-convex}
Let $f(\cdot):\mathbb{R}^d\rightarrow\mathbb{R}$ be a proper closed and convex function, differentiable over $\mathbb{R}^d$. Then, for a given $\mu>0$, the following claims are equivalent:
\begin{itemize}
    \item $f(\cdot)$ is $\mu$-strongly convex.
    \item $f(\boldsymbol{y})\geq f(\boldsymbol{x})+\nabla f(\boldsymbol{x})^T(\boldsymbol{y}-\boldsymbol{x})+\frac{\mu}{2}\|\boldsymbol{y}-\boldsymbol{x}\|^2$ for all $\boldsymbol{x}, \boldsymbol{y}\in\mathbb{R}^d$.
    \item The function $f(\cdot)-\frac{\mu}{2}\|\cdot\|^2$ is convex.
\end{itemize}
\end{lemma}

\section{Proofs for Privacy Analysis}
\label{proof1}

\subsection{Proof of Lemma~\ref{lemma:smooth-reduct}}
\label{proof:smooth-reduct}

\begin{lemma}[Full statement of Lemma~\ref{lemma:smooth-reduct}]
Let $\psi(\cdot)$ and $\psi^\prime(\cdot)$ be two noisy update functions (c.f. (\ref{eq:noise-update})) of DPSGD based on adjacent datasets $\mathcal{D}$ and $\mathcal{D}^\prime$, and $n$ be the size of the dataset $\mathcal{D}$. If the loss function is $L$-smooth (Assumption~\ref{A:smooth-stochastic}), for any random variables $\boldsymbol{\mu}$ and $\boldsymbol{\nu}$, we have
    \begin{equation}
    \mathcal{D}_\alpha^{((1+\eta L)z)}(\mathbb{P}_{\psi(\boldsymbol{\mu})}||\mathbb{P}_{\psi^\prime(\boldsymbol{\nu})})
    \leq \ \mathcal{D}_\alpha^{(z)}(\mathbb{P}_{\boldsymbol{\mu}}||\mathbb{P}_{\boldsymbol{\nu}})+\frac{2\alpha C^2}{\beta nb\sigma_{\textup{DP}}^2}.
    \end{equation}
    If we further assume the mini-batch size $b\leq \frac{n}{5}$, the RDP parameter $\alpha\leq\alpha^*(\frac{b}{n},\frac{b\sqrt{\beta}\sigma_{\textup{DP}}}{2C})$, and the per-dimension Gaussian noise scale $\sigma_{\textup{DP}}>\frac{8C}{b\sqrt{\beta}}$, then
    \begin{equation}
        \mathcal{D}_\alpha^{((1+\eta L)z)}(\mathbb{P}_{\psi(\boldsymbol{\mu})}||\mathbb{P}_{\psi^\prime(\boldsymbol{\nu})}) \leq \ \mathcal{D}_\alpha^{(z)}(\mathbb{P}_{\boldsymbol{\mu}}||\mathbb{P}_{\boldsymbol{\nu}})+\frac{8\alpha C^2}{\beta n^2\sigma_{\textup{DP}}^2}.
    \end{equation}
\end{lemma}

\begin{proof}
    (1) Based on Lemma~\ref{lem:W-distance}, for the shifted R\'enyi divergence $\mathcal{D}_\alpha^{(z)}(\mathbb{P}_{\boldsymbol{\mu}}||\mathbb{P}_{\boldsymbol{\nu}})$, there exist jointly distributed random variables ($\boldsymbol{\mu},\boldsymbol{\mu}^\prime)$ such that $\operatorname{Pr}[||\boldsymbol{\mu}-\boldsymbol{\mu}^\prime||\leq z] = 1$ and $\mathcal{D}_\alpha^{(z)}(\mathbb{P}_{\boldsymbol{\mu}}||\mathbb{P}_{\boldsymbol{\nu}})=\mathcal{D}_\alpha (\mathbb{P}_{\boldsymbol{\mu}^\prime}||\mathbb{P}_{\boldsymbol{\nu}})$. Then, one may obtain
    \begin{equation}
    \|\psi(\boldsymbol{\mu})-\psi(\boldsymbol{\mu}^\prime)\| \leq \|\boldsymbol{\mu}-\boldsymbol{\mu}^\prime\|+\eta L\|\boldsymbol{\mu}-\boldsymbol{\mu}^\prime\| \leq (1+\eta L)z, 
    \end{equation}
    where the first step is by the triangle inequality and the $L$-smooth assumption, and the second step is by the definition of $\boldsymbol{\mu}^\prime$. Thus, we have
    \begin{equation}
        \label{eq:proof31-1}
        \begin{aligned}
        &\mathcal{D}_\alpha^{((1+\eta L)z)}(\mathbb{P}_{\psi(\boldsymbol{\mu})}||\mathbb{P}_{\psi^\prime(\boldsymbol{\nu})}) \\
        &\leq \mathcal{D}_\alpha(\mathbb{P}_{\psi(\boldsymbol{\mu}^\prime)}||\mathbb{P}_{\psi^\prime(\boldsymbol{\nu})}) \\
        &= \mathcal{D}_\alpha(\mathbb{P}_{\psi(\boldsymbol{\mu}^\prime)}||\mathbb{P}_{(1-q)\psi(\boldsymbol{\nu})+q\psi^{\prime\prime}(\boldsymbol{\nu})}) \\
        & \leq (1-q)\mathcal{D}_\alpha(\mathbb{P}_{\psi(\boldsymbol{\mu}^\prime)}||\mathbb{P}_{\psi(\boldsymbol{\nu})})+q\mathcal{D}_\alpha(\mathbb{P}_{\psi(\boldsymbol{\mu}^\prime)}||\mathbb{P}_{\psi^{\prime\prime}(\boldsymbol{\nu})})
        \\
        &\leq (1-q)\mathcal{D}_\alpha(\mathbb{P}_{\boldsymbol{\mu}^\prime}||\mathbb{P}_{\boldsymbol{\nu}})+q\mathcal{D}_\alpha(\mathbb{P}_{\psi(\boldsymbol{\mu}^\prime)}||\mathbb{P}_{\psi^{\prime\prime}(\boldsymbol{\nu})}),
        \end{aligned}
    \end{equation}
    where the first step is by the definition of the shifted R\'enyi divergence, the second step is due to SGD sampling with $\psi^{\prime\prime}(\cdot)\stackrel{\Delta}{=}\psi^{\prime}(\cdot|\bar{\mathcal{D}}\in\mathcal{B}_t)$, $\bar{\mathcal{D}}$ denotes the sole differing entry between adjacent datasets ($\mathcal{D}^\prime=\mathcal{D}\cup \{\bar{\mathcal{D}}\}$), the third step is by Lemma~\ref{lem:convex}, and the last step is by Lemma~\ref{lem:post-process}.
    
    Note that for the second term $\mathcal{D}_\alpha(\mathbb{P}_{\psi(\boldsymbol{\mu}^\prime)}||\mathbb{P}_{\psi^{\prime\prime}(\boldsymbol{\nu})})$, we have
    \begin{equation}
    \begin{aligned}
    &\mathcal{D}_\alpha(\mathbb{P}_{\psi(\boldsymbol{\mu}^\prime)}||\mathbb{P}_{\psi^{\prime\prime}(\boldsymbol{\nu})})\\
    &\leq \mathcal{D}_\alpha(\mathbb{P}_{\psi(\boldsymbol{\mu}^\prime),\boldsymbol{\mu}^\prime}||\mathbb{P}_{\psi^{\prime\prime}(\boldsymbol{\nu}),\boldsymbol{\nu}}) \\
    &\leq \sup_{\boldsymbol{v}}\mathcal{D}_\alpha(\mathbb{P}_{\psi(\boldsymbol{\mu}^\prime)|\boldsymbol{\mu}^\prime=\boldsymbol{v}}||\mathbb{P}_{\psi^{\prime\prime}(\boldsymbol{\nu})|\boldsymbol{\nu}=\boldsymbol{v}})+\mathcal{D}_\alpha(\mathbb{P}_{\boldsymbol{\mu}^\prime}||\mathbb{P}_{\boldsymbol{\nu}}) \\
    &\leq \frac{2\alpha C^2}{\beta b^2\sigma_{\text{DP}}^2} + \mathcal{D}_\alpha(\mathbb{P}_{\boldsymbol{\mu}^\prime}||\mathbb{P}_{\boldsymbol{\nu}}),
    \end{aligned}
    \end{equation}
    where the first step is by Lemma~\ref{lem:post-process}, the second step is by Lemma~\ref{lem:compos}, and the last step is by the well-known result $\mathcal{D}_\alpha(\mathcal{N}(0, \sigma^2 \boldsymbol{I}_d) \| \mathcal{N}(\boldsymbol{u}, \sigma^2 \boldsymbol{I}_d))=\alpha\|\boldsymbol{u}\|_2^2 / 2 \sigma^2$.
    Hence,
    \begin{equation}
    \begin{aligned}
    &\mathcal{D}_\alpha^{((1+\eta L)z)}(\mathbb{P}_{\psi(\boldsymbol{\mu})}||\mathbb{P}_{\psi^\prime(\boldsymbol{\nu})}) \\
    &\leq (1-q)\mathcal{D}_\alpha(\mathbb{P}_{\boldsymbol{\mu}^\prime}||\mathbb{P}_{\boldsymbol{\nu}})+\frac{2\alpha C^2}{\beta nb\sigma_{\text{DP}}^2} + q\mathcal{D}_\alpha(\mathbb{P}_{\boldsymbol{\mu}^\prime}||\mathbb{P}_{\boldsymbol{\nu}})
    \\
    &\leq \mathcal{D}_\alpha^{(z)}(\mathbb{P}_{\boldsymbol{\mu}}||\mathbb{P}_{\boldsymbol{\nu}})+\frac{2\alpha C^2}{\beta nb\sigma_{\text{DP}}^2}.
    \end{aligned}
    \end{equation}

    (2) Assuming $b\leq n/5$, $\alpha\leq\alpha^*(\frac{b}{n},\frac{b\sqrt{\beta}\sigma_{\text{DP}}}{2C})$, and $\sigma_{\text{DP}}>\frac{8C}{b\sqrt{\beta}}$, the proof is similar to the previous case. Starting with (\ref{eq:proof31-1}), we obtain
    \begin{equation}
        \begin{aligned}
        &\mathcal{D}_\alpha^{((1+\eta L)z)}(\mathbb{P}_{\psi(\boldsymbol{\mu})}||\mathbb{P}_{\psi^\prime(\boldsymbol{\nu})}) \\
        &\leq \mathcal{D}_\alpha(\mathbb{P}_{\psi(\boldsymbol{\mu}^\prime),\boldsymbol{\mu}^\prime}||\mathbb{P}_{\psi^\prime(\boldsymbol{\nu}),\boldsymbol{\nu}}) \\
        &\leq\sup_{\boldsymbol{v}}\mathcal{D}_\alpha(\mathbb{P}_{\psi(\boldsymbol{\mu}^\prime)|\boldsymbol{\mu}^\prime=\boldsymbol{v}}||\mathbb{P}_{\psi^{\prime}(\boldsymbol{\nu})|\boldsymbol{\nu}=\boldsymbol{v}}) +\mathcal{D}_\alpha(\mathbb{P}_{\boldsymbol{\mu}^\prime}||\mathbb{P}_{\boldsymbol{\nu}}) \\
        & \leq \mathcal{D}_\alpha^{\text{SGM}}\left(\frac{b}{n},\frac{ b\sqrt{\beta}\sigma_{\text{DP}}}{2C}\right)+\mathcal{D}_\alpha(\mathbb{P}_{\boldsymbol{\mu}^\prime}||\mathbb{P}_{\boldsymbol{\nu}})\\
        &\leq \mathcal{D}_\alpha^{(z)}(\mathbb{P}_{\boldsymbol{\mu}}||\mathbb{P}_{\boldsymbol{\nu}})+\frac{8\alpha C^2}{\beta n^2\sigma_{\text{DP}}^2},
        \end{aligned}
    \end{equation}
    where the first step is by Lemma~\ref{lem:post-process}, the second step is by Lemma~\ref{lem:compos}, the last step is by Lemma~\ref{lem:SGM}.
\end{proof}

\subsection{Proof of Theorem~\ref{thm:privacy-GC}}
\label{proof:privacy-GC}

\begin{theorem}[Full statement of Theorem~\ref{thm:privacy-GC}]
Given the number of total SGD  iterations $T$, dataset size $n$, batch size $b$, stepsize $\eta$, $\alpha> 1$, gradient clipping threshold $C$, and noise scale $\sigma_{\textup{DP}}$, if the loss function is $L$-smooth (Assumption~\ref{A:smooth-stochastic}), then the DPSGD-GC algorithm satisfies $(\alpha,\varepsilon)$-RDP for
\begin{equation}
    \varepsilon = \mathcal{O}\left(\frac{\alpha C^2}{nb\sigma_{\textup{DP}}^2}T\right).
\end{equation}
If we further assume the mini-batch size $b\leq \frac{n}{5}$, RDP parameter $\alpha\leq\alpha^*(\frac{b}{n},\frac{b\sqrt{\beta}\sigma_{\textup{DP}}}{2C})$, and the per-dimension Gaussian noise scale $\sigma_{\textup{DP}}>\frac{8C}{b\sqrt{\beta}}$ with a constant $\beta\in(0,1)$, then
\begin{equation}
    \varepsilon = \mathcal{O}\left(\frac{\alpha C^2}{ n^2\sigma_{\textup{DP}}^2}T\right).
\end{equation}
\end{theorem}

\begin{proof}
    We first rewrite the update procedure of DPSGD-GC as follows:
    \begin{equation}
    \boldsymbol{\theta}_{t+1} = \boldsymbol{\theta}_t-\frac{\eta}{b}\sum_{\xi \in \mathcal{B}_t} \operatorname{clip}\left(\nabla l_\xi\left(\boldsymbol{\theta}_t\right)\right)+\boldsymbol{\varrho}_t+\boldsymbol{\varsigma}_t  \stackrel{\Delta}{=} \psi(\boldsymbol{\theta}_t) + \boldsymbol{\varsigma}_t,
    \end{equation}
    where $\psi(\cdot)$ denotes the noisy update function, and $\boldsymbol{\varrho}_t\sim\mathcal{N}(\beta\eta^2\sigma_{\text{DP}}^2\boldsymbol{I}_d)$ and $\boldsymbol{\varsigma}_t\sim\mathcal{N}((1-\beta)\eta^2\sigma_{\text{DP}}^2\boldsymbol{I}_d)$ are both the zero-mean Gaussian perturbation with $\beta\in(0,1)$.

    Then, consider real sequence $\{a_t\}_{i=0}^{T-1}$ such that $z_t=\sum_{i=0}^{t-1}(1+\eta L)^{t-i-1}(-a_i)$ is non-negative for all $t$ and $z_T=0$. By this way, we have $z_0=0$ and $z_{t+1}=(1+\eta L)z_t -a_t$.
    The proof proceeds by induction, leveraging Lemma~\ref{lemma:shifted-reduct} and \ref{lemma:smooth-reduct}. Specifically, we have
    \begin{equation}
    \begin{aligned}
    &\mathcal{D}_\alpha^{(z_{t+1})}(\mathbb{P}_{\boldsymbol{\theta}_{t+1}}||\mathbb{P}_{\boldsymbol{\theta}^\prime_{t+1}})\\
    &= \mathcal{D}_\alpha^{(z_{t+1})}(\mathbb{P}_{\psi(\boldsymbol{\theta}_{t})+\boldsymbol{\varsigma}_t}||\mathbb{P}_{\psi^\prime(\boldsymbol{\theta}^\prime_{t})+\boldsymbol{\varsigma}_t})\\
    &\leq \mathcal{D}_\alpha^{(z_{t+1}+a_{t})}(\mathbb{P}_{\psi(\boldsymbol{\theta}_t)}||\mathbb{P}_{\psi^\prime(\boldsymbol{\theta}^\prime_{t})}) + \frac{\alpha a_t^2}{2\eta^2\sigma_{\text{DP}}^2(1-\beta)}\\
    &= \mathcal{D}_\alpha^{((1+\eta L)z_t)}(\mathbb{P}_{\psi(\boldsymbol{\theta}_t)}||\mathbb{P}_{\psi^\prime(\boldsymbol{\theta}^\prime_{t})}) + \frac{\alpha a_t^2}{2\eta^2\sigma_{\text{DP}}^2(1-\beta)}\\
    & \leq \mathcal{D}^{(z_t)}_\alpha(\mathbb{P}_{\boldsymbol{\theta}_t}||\mathbb{P}_{\boldsymbol{\theta}_t^\prime}) + \frac{2\alpha C^2}{\beta nb\sigma_{\text{DP}}^2} + \frac{\alpha a_t^2}{2\eta^2\sigma_{\text{DP}}^2(1-\beta)},
    \end{aligned}
    \end{equation}
    where the second step is by Lemma~\ref{lemma:shifted-reduct}, the third step is by the definition of $z_{t+1}$, and the last step is by Lemma~\ref{lemma:smooth-reduct}. By using the induction hypothesis, we have
    \begin{equation}
    \begin{aligned}
    &\mathcal{D}_{\alpha}(\mathbb{P}_{\boldsymbol{\theta}_{T}}||\mathbb{P}_{\boldsymbol{\theta}^\prime_{T}})\\
    &\leq \mathcal{D}_\alpha^{(z_0)}(\mathbb{P}_{\boldsymbol{\theta}_0}||\mathbb{P}_{\boldsymbol{\theta}_0^\prime})+\sum_{t=0}^{T-1}\frac{\alpha a_t^2}{2\eta^2\sigma_{\text{DP}}^2(1-\beta)} + \sum_{t=0}^{T-1}\frac{2\alpha C^2}{\beta nb\sigma_{\text{DP}}^2} \\
    &= \sum_{t=0}^{T-1}\frac{\alpha a_t^2}{2\eta^2\sigma_{\text{DP}}^2(1-\beta)} + \sum_{t=0}^{T-1}\frac{2\alpha C^2}{\beta nb\sigma_{\text{DP}}^2}.
    \end{aligned}
    \end{equation}  
    Let $a_t=0$ for all $t$, we have
    \begin{equation}
    \label{eq-linear}
    \mathcal{D}_{\alpha}(\mathbb{P}_{\boldsymbol{\theta}_{T}}||\mathbb{P}_{\boldsymbol{\theta}^\prime_{T}})\leq \frac{2\alpha C^2}{\beta nb\sigma_{\text{DP}}^2}T.
    \end{equation}
    Under additional parameter constraints, the proof follows directly by applying the corresponding version of Lemma~\ref{lemma:smooth-reduct}. Consequently, we omit the detailed derivations here.
\end{proof}

\subsection{Proof of Theorem~\ref{thm:privacy-DC}}
\label{proof:privacy-DC}

\begin{theorem}[(Full statement of Theorem~\ref{thm:privacy-DC}]
Given the number of total SGD iterations $T$, dataset size $n$, batch size $b$, stepsize $\eta$, $\alpha>1$, gradient clipping threshold $C$, diameter of the model-parameter domain $D$, and noise scale $\sigma_{\textup{DP}}$, if the loss function is $L$-smooth (Assumption~\ref{A:smooth-stochastic}), then the DPSGD-DC algorithm satisfies $(\alpha,\varepsilon)$-RDP for
\begin{equation}
    \varepsilon=\mathcal{O}\left(\frac{\alpha C^2}{nb\sigma_{\textup{DP}}^2}\min\left\{T, \frac{(1+\eta L)^2nbD^2}{\eta^2C^2}\right\}\right).
\end{equation}
If we further assume the mini-batch size $b\leq \frac{n}{5}$, RDP parameter $\alpha\leq\alpha^*(\frac{b}{n},\frac{b\sqrt{\beta}\sigma_{\textup{DP}}}{2C})$, and the per-dimension Gaussian noise scale $\sigma_{\textup{DP}}>\frac{8C}{b\sqrt{\beta}}$ with a constant $\beta\in(0,1)$, then
\begin{equation}
    \varepsilon=\mathcal{O}\left(\frac{\alpha C^2}{ n^2\sigma_{\textup{DP}}^2}\min\left\{T, \frac{(1+\eta L)^2n^2D^2}{\eta^2C^2}\right\}\right).
\end{equation}
\end{theorem}

\begin{proof}
    Similarly, consider the real sequence $\{a_t\}_{i=0}^{T-1}$ and any $\tau\in \{0,1,\cdots,T-1\}$ such that $z_t=(1+\eta L)^{t-\tau}D+\sum_{i=\tau}^{t-1}(1+\eta L)^{t-i-1}(-a_i)$ is non-negative for all $t\geq\tau$ and $z_T=0$. Note that we also have $ z_{t+1}=(1+\eta L)z_t -a_t$, yielding
    \begin{equation}
    \begin{aligned}
    &\mathcal{D}_\alpha^{(z_{t+1})}(\mathbb{P}_{\boldsymbol{\theta}_{t+1}}||\mathbb{P}_{\boldsymbol{\theta}^\prime_{t+1}})\\
    &\leq \mathcal{D}^{(z_t)}_\alpha(\mathbb{P}_{\boldsymbol{\theta}_t}||\mathbb{P}_{\boldsymbol{\theta}_t^\prime}) + \frac{2\alpha C^2}{\beta nb\sigma_{\text{DP}}^2} + \frac{\alpha a_t^2}{2\eta^2\sigma_{\text{DP}}^2(1-\beta)}.
    \end{aligned}
    \end{equation}
    By repeating the induction from $T$ to $\tau$, we can obtain
    \begin{equation}
    \begin{aligned}
    &\mathcal{D}_{\alpha}(\mathbb{P}_{\boldsymbol{\theta}_{T}}||\mathbb{P}_{\boldsymbol{\theta}^\prime_{T}})\\
    &\leq \mathcal{D}_\alpha^{(z_\tau)}(\mathbb{P}_{\boldsymbol{\theta}_\tau}||\mathbb{P}_{\boldsymbol{\theta}_\tau^\prime})+\sum_{t=\tau}^{T-1}\frac{\alpha a_t^2}{2\eta^2\sigma_{\text{DP}}^2(1-\beta)} + \sum_{t=\tau}^{T-1}\frac{2\alpha C^2}{\beta nb\sigma_{\text{DP}}^2} \\
    &= \sum_{t=\tau}^{T-1}\frac{\alpha a_t^2}{2\eta^2\sigma_{\text{DP}}^2(1-\beta)} + \sum_{t=\tau}^{T-1}\frac{2\alpha C^2}{\beta nb\sigma_{\text{DP}}^2},
    \end{aligned}
    \end{equation}
    where the last step is by $z_\tau=D$. Let $a_\tau=(1+\eta L)D$, $a_t=0$ for all $t>\tau$, we have 
    \begin{equation}
    \begin{aligned}
    &\mathcal{D}_{\alpha}(\mathbb{P}_{\boldsymbol{\theta}_{T}}||\mathbb{P}_{\boldsymbol{\theta}^\prime_{T}})\\
    &\leq \min_{\tau\in\{0,\cdots,T-1\}} \frac{2\alpha C^2}{\beta nb\sigma_{\text{DP}}^2}(T-\tau)+\frac{\alpha (1+\eta L)^2D^2}{2\eta^2\sigma_{\text{DP}}^2(1-\beta)} \\
    &=\frac{2\alpha C^2}{\beta nb\sigma_{\text{DP}}^2}+\frac{\alpha (1+\eta L)^2D^2}{2\eta^2\sigma_{\text{DP}}^2(1-\beta)}.
    \end{aligned}
    \end{equation}
    Define 
    \begin{equation}
        g(\beta)=\frac{2\alpha C^2}{\beta nb\sigma_{\text{DP}}^2}+\frac{\alpha (1+\eta L)^2D^2}{2\eta^2\sigma_{\text{DP}}^2(1-\beta)}.
    \end{equation}
    By setting $g^\prime(\beta)=0$, one may obtain
    \begin{equation}
        \beta^* = \frac{\sqrt{\frac{2\alpha C^2}{nb\sigma_{\text{DP}}^2}}}{\sqrt{\frac{2\alpha C^2}{nb\sigma_{\text{DP}}^2}}+\sqrt{\frac{\alpha (1+\eta L)^2D^2}{2\eta^2\sigma_{\text{DP}}^2}}}.
    \end{equation}
    Substituting $\beta^*$ into the original optimization problems, it becomes
    \begin{equation}
        g(\beta^*) = \left(\sqrt{\frac{2\alpha C^2}{nb\sigma_{\text{DP}}^2}}+\sqrt{\frac{\alpha (1+\eta L)^2D^2}{2\eta^2\sigma_{\text{DP}}^2}}\right)^2.
    \end{equation}
    Combined with the linear bound in (\ref{eq-linear}), we can obtain
    \begin{equation}
        \mathcal{D}_{\alpha}(\mathbb{P}_{\boldsymbol{\theta}_{T}}||\mathbb{P}_{\boldsymbol{\theta}^\prime_{T}})\!\leq\!\mathcal{O}\left(\frac{\alpha C^2}{ nb\sigma_{\textup{DP}}^2}\!\min\left\{T, \frac{(1+\eta L)^2nbD^2}{2\eta^2C^2}\right\}\right).
    \end{equation}
    The proof under additional assumptions is also omitted here.
\end{proof}

\section{Proofs for Utility Analysis}
\label{proof2}

Our utility analysis builds upon the results of \citet{bassily2014private}, \citet{zhang2020improved}, and \citet{koloskova2023revisiting}.

\subsection{Proof of Theorem~\ref{thm:converge-DC}}
\label{proof:converge-DC}

\begin{proof} 
Recall that the update procedure of DPSGD-DC is as follows:
\begin{equation}
\begin{aligned}
\boldsymbol{\theta}_{t+1} &= \Pi_\mathcal{K}\left(\boldsymbol{\theta}_t-\eta g(\boldsymbol{\theta}_t)+\boldsymbol{\zeta}_t\right), 
\end{aligned}
\end{equation}
where 
\begin{equation}
    g(\boldsymbol{\theta}_t)=\frac{1}{b}\sum_{\xi \in \mathcal{B}_t} g_\xi(\boldsymbol{\theta}_t)=\frac{1}{b}\sum_{\xi \in \mathcal{B}_t} \operatorname{clip}_C\left(\nabla l_\xi(\boldsymbol{\theta}_t)\right),    
\end{equation}
and $\boldsymbol{\zeta}_t\sim\mathcal{N}(\boldsymbol{0}, \eta^2\sigma_{\text{DP}}^2\boldsymbol{I}_d)$ is the Gaussian perturbation.

We now proceed to the proof of Theorem~\ref{thm:converge-DC}. It is worth noting that the main challenges lie in the gradient clipping operation, the SGD procedure, and the parameter projection step. Therefore, the proof will be divided into cases to address these aspects separately.

(1) $C\leq10\sigma_{\text{SGD}} (\frac{L}{\mu})^{\frac{1}{2}}$

(1.1) $\|\nabla l(\boldsymbol{\theta}_t)\|\geq 35\sigma_{\text{SGD}} (\frac{L}{\mu})^{\frac{3}{4}}$

We start the analysis with the following inequality:
\begin{equation}
\label{eq:converge-1}
\begin{aligned}
&\mathbb{E}_{\boldsymbol{\zeta}_t}\left[\|\boldsymbol{\theta}_{t+1}-\boldsymbol{\theta}^*\|^2\right] \\
&\leq \mathbb{E}_{\boldsymbol{\zeta}_t}\left[\|\boldsymbol{\theta}_t-\eta g(\boldsymbol{\theta}_t)+\boldsymbol{\zeta}_t-\boldsymbol{\theta}^*\|^2\right] \\
&\leq \|\boldsymbol{\theta}_t-\eta g(\boldsymbol{\theta}_t)-\boldsymbol{\theta}^*\|^2+d\eta^2\sigma_{\text{DP}}^2 \\
&\leq \|\boldsymbol{\theta}_t-\boldsymbol{\theta}^*\|^2-\frac{1}{b}\sum_{\xi\in\mathcal{B}_t}2\eta [g_\xi(\boldsymbol{\theta}_t)]^T(\boldsymbol{\theta}_t-\boldsymbol{\theta}^*)\\
&+\eta^2C^2+d\eta
^2 \sigma_{\text{DP}}^2,
\end{aligned}
\end{equation}
where the first step is by the independence of noise and Lemma~\ref{lem:project}. Let $\gamma_\xi=\min (1, \frac{C}{\|\nabla l_\xi(\boldsymbol{\theta}_t)\|})$. If $\|\nabla l_\xi(\boldsymbol{\boldsymbol{\theta}}_t)-\nabla l(\boldsymbol{\theta}_t)\|\leq 5\sigma_{\text{SGD}} (\frac{L}{\mu})^{\frac{1}{4}}$, we immediate get $\gamma_\xi \leq \frac{C}{35\sigma_{\text{SGD}} (\frac{L}{\mu})^{3/4}-5\sigma_{\text{SGD}} (\frac{L}{\mu})^{1/4}}$ and $\gamma_\xi\geq \frac{7C}{8\|\nabla l(\boldsymbol{\theta}_t)\|}$. Thus,
\begin{equation}
\begin{aligned}
&-2\eta [g_\xi(\boldsymbol{\theta}_t)]^T(\boldsymbol{\theta}_t-\boldsymbol{\theta}^*)\\
&=-2\eta[g_\xi(\boldsymbol{\theta}_t)-\gamma_\xi\nabla l(\boldsymbol{\theta}_t)+\gamma_\xi\nabla l(\boldsymbol{\theta}_t)]^T(\boldsymbol{\theta}_t-\boldsymbol{\theta}^*) \\
&\leq-2\eta\gamma_\xi [\nabla l(\boldsymbol{\theta}_t)]^T(\boldsymbol{\theta}_t-\boldsymbol{\theta}^*)\\
&-2\eta[g_\xi(\boldsymbol{\theta}_t)-\gamma_\xi\nabla l(\boldsymbol{\theta}_t)]^T(\boldsymbol{\theta}_t-\boldsymbol{\theta}^*) \\
&\leq - \frac{7\eta C}{4\|\nabla l(\boldsymbol{\theta}_t)\|}[l(\boldsymbol{\theta}_t)-l(\boldsymbol{\theta}^*)]\\
&+2\eta \gamma_\xi \|\nabla l_\xi(\boldsymbol{\theta}_t)-\nabla l(\boldsymbol{\theta}_t)\| \cdot \|\boldsymbol{\theta}_t-\boldsymbol{\theta}^*\| \\
&\leq -\frac{7\eta C}{4\sqrt{2L}}\sqrt{l(\boldsymbol{\theta}_t)-l(\boldsymbol{\theta}^*)}\\
&+\frac{2\eta C\sigma_{\text{SGD}} (\frac{L}{\mu})^{\frac{1}{4}}}{7\sigma_{\text{SGD}} (\frac{L}{\mu})^{\frac{3}{4}}-\sigma_{\text{SGD}} (\frac{L}{\mu})^{\frac{1}{4}}}\sqrt{\frac{2}{\mu}}\sqrt{l(\boldsymbol{\theta}_t)-l(\boldsymbol{\theta}^*)} \\
&\leq-\frac{7\eta C}{4\sqrt{2L}}\sqrt{l(\boldsymbol{\theta}_t)-l(\boldsymbol{\theta}^*)}\\
&+\frac{4\eta C}{\sqrt{2}(7\sqrt{L}-\sqrt{\mu})}\sqrt{l(\boldsymbol{\theta}_t)-l(\boldsymbol{\theta}^*)} \\
&\leq -\frac{13\eta C}{12\sqrt{2L}}\sqrt{l(\boldsymbol{\theta}_t)-l(\boldsymbol{\theta}^*)},
\end{aligned}
\end{equation}
where the third step is by the convexity and Cauchy–Schwarz inequality, and the fourth step is by Lemma~\ref{lem:smooth} and \ref{lem:strong-convex}.

Else if $\|\nabla l_\xi(\boldsymbol{\theta}_t)-\nabla l(\boldsymbol{\theta}_t)\|> 5\sigma_{\text{SGD}} (\frac{L}{\mu})^{\frac{1}{4}}$, we have
\begin{equation}
\begin{aligned}
-2\eta [g_\xi(\boldsymbol{\theta}_t)]^T(\boldsymbol{\theta}_t-\boldsymbol{\theta}^*) &\leq 2\eta C\|\boldsymbol{\theta}_t-\boldsymbol{\theta}^*\| \\
&\leq \frac{4\eta C}{\sqrt{2\mu}}\sqrt{l(\boldsymbol{\theta}_t)-l(\boldsymbol{\theta}^*)},
\end{aligned}
\end{equation}
by using Cauchy–Schwarz inequality and Lemma~\ref{lem:strong-convex}. Here, we define $\kappa_\xi=\boldsymbol{1}\{\|\nabla l_\xi(\boldsymbol{\theta}_t)-\nabla l(\boldsymbol{\theta}_t)\|> 5\sigma_{\text{SGD}} (\frac{L}{\mu})^{\frac{1}{4}}\}$. According to Lemma~\ref{lem:markov}, we have
\begin{equation}
\begin{aligned}
\operatorname{Pr}(\kappa_\xi=1)&\!=\!\operatorname{Pr}\left(\|\nabla l_\xi(\boldsymbol{\theta}_t)-\nabla l(\boldsymbol{\theta}_t)\|^2> 25\sigma_{\text{SGD}}^2 \sqrt{L/\mu}\right)\\
&\leq\frac{\sigma_{\text{SGD}}^2}{25\sigma_{\text{SGD}}^2\sqrt{L/\mu}}=\frac{1}{25}\sqrt{\frac{\mu}{L}}, \\
\operatorname{Pr}(\kappa_\xi=0)&\geq 1-\frac{1}{25}\sqrt{\frac{\mu}{L}}\geq \frac{24}{25}.
\end{aligned}
\end{equation}
Hence,
\begin{equation}
\label{eq:converge-2}
\begin{aligned}
&\mathbb{E}_\xi\left[-2\eta [g_\xi(\boldsymbol{\theta}_t)]^T(\boldsymbol{\theta}_t-\boldsymbol{\theta}^*)\right]\\
&\leq -\frac{13\eta C}{12\sqrt{2L}}\sqrt{l(\boldsymbol{\theta}_t)-l(\boldsymbol{\theta}^*)}\cdot \frac{24}{25}\\
&+\frac{4\eta C}{\sqrt{2\mu}}\sqrt{l(\boldsymbol{\theta}_t)-l(\boldsymbol{\theta}^*)}\cdot \frac{1}{25}\sqrt{\frac{\mu}{L}}\\
&\leq -\frac{22\eta C}{25\sqrt{2L}}\sqrt{l(\boldsymbol{\theta}_t)-l(\boldsymbol{\theta}^*)}.
\end{aligned}
\end{equation}
Then, we choose $\eta\leq\frac{7}{10L}(\frac{L}{\mu})^{1/4}$, yielding
\begin{equation}
\label{eq:converge-3}
\begin{aligned}
\eta^2C^2 &\leq \frac{7\eta C^2}{10L}(\frac{L}{\mu})^{1/4}\leq\frac{\eta C^2}{2L} \frac{4\|\nabla l(\boldsymbol{\theta}_t)\|}{10C} \\
&\leq \frac{\eta C}{\sqrt{2L}}\frac{2\sqrt{l(\boldsymbol{\theta}_t)-l(\boldsymbol{\theta}^*)}}{5},
\end{aligned}
\end{equation}
where the last step is by Lemma~\ref{lem:smooth}. Substituting (\ref{eq:converge-2}) and (\ref{eq:converge-3}) into (\ref{eq:converge-1}), we have
\begin{equation}
\begin{aligned}
&\mathbb{E}\left[\|\boldsymbol{\theta}_{t+1}-\boldsymbol{\theta}^*\|^2\right] \\
&\leq \|\boldsymbol{\theta}_t-\boldsymbol{\theta}^*\|^2-\frac{22\eta c}{25\sqrt{2L}}\sqrt{l(\boldsymbol{\theta}_t)-l(\boldsymbol{\theta}^*)}\\
&+\frac{\eta c}{\sqrt{2L}}\frac{2\sqrt{l(\boldsymbol{\theta}_t)-l(\boldsymbol{\theta}^*)}}{5}+d\eta
^2 \sigma_{\text{DP}}^2 \\
&\leq \|\boldsymbol{\theta}_t-\boldsymbol{\theta}^*\|^2-\frac{12\eta c}{25\sqrt{2L}}\sqrt{l(\boldsymbol{\theta}_t)-l(\boldsymbol{\theta}^*)}+d\eta^2\sigma_{\text{DP}}^2.
\end{aligned}
\end{equation}
Thus, averaging over $t$, we have
\begin{equation}
\begin{aligned}
&\frac{1}{T+1}\sum_{t=0}^T \mathbb{E}\left[\sqrt{l(\boldsymbol{\theta}_t)-l(\boldsymbol{\theta}^*)}\right]\\
&\leq \mathcal{O}\left(\frac{\sqrt{L}D^2}{\eta C T}+\frac{d\eta\sigma_{\text{DP}}^2\sqrt{L}}{C}\right).
\end{aligned}
\end{equation}

(1.2) $\|\nabla l(\boldsymbol{\theta}_t)\|< 35\sigma_{\text{SGD}} (\frac{L}{\mu})^{\frac{3}{4}}$

By using Lemma~\ref{lem:strong-convex}, we immediate obtain 
\begin{equation}
    \sqrt{l(\boldsymbol{\theta}_t)-l(\boldsymbol{\theta}^*)}\leq \sqrt{\frac{1}{2\mu}}\|\nabla l(\boldsymbol{\theta}_t)\|\leq \mathcal{O}\left(\frac{L^{3/4}}{\mu^{5/4}}\sigma_{\text{SGD}}\right).
\end{equation}

(2) $C\geq 10\sigma_{\text{SGD}} (\frac{L}{\mu})^{\frac{1}{2}}$

(2.1)  $\|\nabla l(\boldsymbol{\theta}_t)\|>\frac{C}{2}$

In this case, we start the analysis with the following inequality:
\begin{equation}
\begin{aligned}
\label{eq:converge-4}
&\mathbb{E}_{\boldsymbol{\zeta}_t}\left[\|\boldsymbol{\theta}_{t+1}-\boldsymbol{\theta}^*\|^2\right] \\
&\leq \mathbb{E}_{\zeta_t}\left[\|\boldsymbol{\theta}_t-\eta g(\boldsymbol{\theta}_t)+\boldsymbol{\zeta}_t-\boldsymbol{\theta}^*\|^2\right] \\
&\leq \|\boldsymbol{\theta}_t-\boldsymbol{\theta}^*\|^2-\frac{1}{b}\sum_{\xi\in \mathcal{B}_t}2\eta [g_\xi(\boldsymbol{\theta}_t)]^T(\boldsymbol{\theta}_t-\boldsymbol{\theta}^*)\\
&+\eta^2\|g(\boldsymbol{\theta}_t)\|^2+d\eta
^2 \sigma_{\text{DP}}^2,
\end{aligned}
\end{equation}
where the first step is by Lemma~\ref{lem:project}, and the second step is by the independence of noise.
Let $\gamma=\min (1, \frac{C}{\|\nabla l(\boldsymbol{\theta}_t)\|})$, we then have $\gamma\geq\min(1,\frac{C}{2\|\nabla l(\boldsymbol{\theta}_t)\|})=\frac{C}{2\|\nabla l(\boldsymbol{\theta}_t)\|}$. Thus,
\begin{equation}
\begin{aligned}
&\mathbb{E}_\xi\left[-2\eta [g_\xi(\boldsymbol{\theta}_t)]^T(\boldsymbol{\theta}_t-\boldsymbol{\theta}^*)\right]\\
&=\mathbb{E}_\xi\left[-2\eta[g_\xi(\boldsymbol{\theta}_t)-\gamma\nabla l(\boldsymbol{\theta}_t)+\gamma\nabla l(\boldsymbol{\theta}_t)]^T(\boldsymbol{\theta}_t-\boldsymbol{\theta}^*)\right] \\
&\leq-2\eta\gamma [\nabla l(\boldsymbol{\theta}_t)]^T(\boldsymbol{\theta}_t-\boldsymbol{\theta}^*)\\
&-\mathbb{E}_\xi\left[2\eta[g_\xi(\boldsymbol{\theta}_t)-\gamma\nabla l(\boldsymbol{\theta}_t)]^T(\boldsymbol{\theta}_t-\boldsymbol{\theta}^*)\right] \\
&\leq -\frac{\eta C}{\|\nabla l(\boldsymbol{\theta}_t)\|}[l(\boldsymbol{\theta}_t)-l(\boldsymbol{\theta}^*)]\\
&+\mathbb{E}_\xi\left[2\eta \|\nabla l_\xi(\boldsymbol{\theta}_t)-\nabla l(\boldsymbol{\theta}_t)\| \cdot \|\boldsymbol{\theta}_t-\boldsymbol{\theta}^*\|\right] \\
&\leq -\frac{\eta C}{\sqrt{2L}}\sqrt{l(\boldsymbol{\theta}_t)-l(\boldsymbol{\theta}^*)}+2\eta \sigma_{\text{SGD}} \sqrt{\frac{2}{\mu}}\sqrt{l(\boldsymbol{\theta}_t)-l(\boldsymbol{\theta}^*)} \\
&\leq-\frac{\eta C}{\sqrt{2L}}\sqrt{l(\boldsymbol{\theta}_t)-l(\boldsymbol{\theta}^*)}+ \frac{2\eta C}{5\sqrt{2L}}\sqrt{l(\boldsymbol{\theta}_t)-l(\boldsymbol{\theta}^*)}\\
&\leq -\frac{3\eta C}{5\sqrt{2L}}\sqrt{l(\boldsymbol{\theta}_t)-l(\boldsymbol{\theta}^*)},
\end{aligned}
\end{equation}
where the third step is by the convexity and Cauchy-Schwarz inequality, and the fourth step is by Lemma~\ref{lem:smooth} and \ref{lem:strong-convex}. Combined with (\ref{eq:converge-3}), we have
\begin{equation}
\begin{aligned}
&\mathbb{E}\left[\|\boldsymbol{\theta}_{t+1}-\boldsymbol{\theta}^*\|^2\right] \\
&\leq \|\boldsymbol{\theta}_t-\boldsymbol{\theta}^*\|^2-\frac{3\eta C}{5\sqrt{2L}}\sqrt{l(\boldsymbol{\theta}_t)-l(\boldsymbol{\theta}^*)}\\
&+\frac{\eta C}{\sqrt{2L}}\frac{2\sqrt{l(\boldsymbol{\theta}_t)-l(\boldsymbol{\theta}^*)}}{5}+\eta
^2 \sigma_{\text{DP}}^2 \\
&\leq \|\boldsymbol{\theta}_t-\boldsymbol{\theta}^*\|^2-\frac{\eta C}{5\sqrt{2L}}\sqrt{l(\boldsymbol{\theta}_t)-l(\boldsymbol{\theta}^*)}+d\eta^2\sigma_{\text{DP}}^2.
\end{aligned}
\end{equation}
Thus, averaging over $t$, we have
\begin{equation}
\begin{aligned}
&\frac{1}{T+1}\sum_{t=0}^T \mathbb{E}\left[\sqrt{l(\boldsymbol{\theta}_t)-l(\boldsymbol{\theta}^*)}\right]\\
&\leq \mathcal{O}\left(\frac{\sqrt{L}D^2}{\eta C T}+\frac{d\eta\sigma_{\text{DP}}^2\sqrt{L}}{C}\right).
\end{aligned}
\end{equation}

(2.2) $\|\nabla l(\boldsymbol{\theta}_t)\|\leq \frac{C}{2}$

In this case, we employ the triangle inequality and obtain
\begin{equation}
\|\nabla l_\xi(\boldsymbol{\theta}_t)\|\leq\|\nabla l_\xi(\boldsymbol{\theta}_t)-\nabla l(\boldsymbol{\theta}_t)\|+\frac{C}{2}.
\end{equation}
Let $\kappa=\boldsymbol{1}\{\|\nabla l_\xi(\boldsymbol{\theta}_t)\|> C\}$ and use Lemma~\ref{lem:markov}, yielding
\begin{equation}
\begin{aligned}
\operatorname{Pr}\left[\kappa=1\right]&\leq \operatorname{Pr}\left[\{\|\nabla l_\xi(\boldsymbol{\theta}_t)-\nabla l(\boldsymbol{\theta}_t)\|^2> \frac{C^2}{4}\}\right]\\
&\leq \frac{4\sigma_{\text{SGD}}^2}{C}.
\end{aligned}
\end{equation}

Hence,
\begin{equation}
\label{eq:converge-5}
\begin{aligned}
&\mathbb{E}_\xi\left[-2\eta [g_\xi(\boldsymbol{\theta}_t)]^T(\boldsymbol{\theta}_t-\boldsymbol{\theta}^*)\right]\\
&=-2\eta \left[\mathbb{E}_\xi[g_\xi(\boldsymbol{\theta}_t)]-\nabla l(\boldsymbol{\theta}_t)+\nabla l(\boldsymbol{\theta}_t)\right]^T(\boldsymbol{\theta}_t-\boldsymbol{\theta}^*) \\
&\leq 2\eta \cdot\|\mathbb{E}_\xi[g_\xi(\boldsymbol{\theta}_t)]-\nabla l(\boldsymbol{\theta}_t)\| \cdot\|\boldsymbol{\theta}_t-\boldsymbol{\theta}^*\|\\
&-2\eta \nabla l(\boldsymbol{\theta}_t)^T(\boldsymbol{\theta}_t-\boldsymbol{\theta}^*) \\
&\leq 2\eta\cdot\|\mathbb{E}_\xi\left[g_\xi(\boldsymbol{\theta}_t)-\nabla l_\xi(\boldsymbol{\theta}_t)|\kappa=1\right]\cdot\operatorname{Pr}(\kappa=1)\|\\
&\cdot \|\boldsymbol{\theta}_t-\boldsymbol{\theta}^*\|-2\eta [l(\boldsymbol{\theta}_t)-l(\boldsymbol{\theta}^*)] \\
&\leq \frac{8\eta\sigma_{\text{SGD}}^2}{C^2}\|\mathbb{E}_\xi[(1-\frac{C}{\|\nabla l_\xi(\boldsymbol{\theta}_t)\|})\nabla l_\xi(\boldsymbol{\theta}_t)]\|\cdot\|\boldsymbol{\theta}_t-\boldsymbol{\theta}^*\|\\
&-2\eta [l(\boldsymbol{\theta}_t)-l(\boldsymbol{\theta}^*)] \\
&\leq \frac{8\eta\sigma_{\text{SGD}}^2}{C^2} \mathbb{E}_\xi[||\nabla l_\xi(\boldsymbol{\theta}_t)-\nabla l(\boldsymbol{\theta}_t)+\nabla l(\boldsymbol{\theta}_t)||]\cdot\|\boldsymbol{\theta}_t-\boldsymbol{\theta}^*\|\\
&-2\eta [l(\boldsymbol{\theta}_t)-l(\boldsymbol{\theta}^*)] \\
&\leq \frac{4\eta\sigma_{\text{SGD}}^3}{\mu C} + \frac{16\eta\sigma_{\text{SGD}}^2}{C^2}\sqrt{\frac{L}{\mu}} [l(\boldsymbol{\theta}_t)-l(\boldsymbol{\theta}^*)]\\
&-2\eta [l(\boldsymbol{\theta}_t)-l(\boldsymbol{\theta}^*)] \\
&\leq \frac{4\eta\sigma_{\text{SGD}}^3}{\mu C} - \left(2\eta-\frac{4}{25}\eta\sqrt{\frac{\mu}{L}}\right)[l(\boldsymbol{\theta}_t)-l(\boldsymbol{\theta}^*)] \\
&\leq \frac{4\eta\sigma_{\text{SGD}}^3}{\mu C} - \frac{46}{25}\eta[l(\boldsymbol{\theta}_t)-l(\boldsymbol{\theta}^*)],
\end{aligned}
\end{equation}
where the second step is by the convexity and Cauchy-Schwarz inequality, the third step is by the law of total expectation, the fifth step is by Jensen's inequality, and the third to last step is by Lemma~\ref{lem:smooth} and \ref{lem:strong-convex}.

Further, note that when $\eta\leq\frac{9}{20L}$,
\begin{equation}
\label{eq:converge-6}
\begin{aligned}
&\mathbb{E}\left[\eta^2\|g(\boldsymbol{\theta}_t)\|^2\right]\\
&\leq \frac{\eta^2}{b^2}\sum_{\xi\in\mathcal{B}_t}\mathbb{E}_\xi\left[\|g_\xi(\boldsymbol{\theta}_t)-\nabla l(\boldsymbol{\theta}_t)+l(\boldsymbol{\theta}_t)\|^2\right] \\
&\leq \frac{2\eta^2}{b^2}\sum_{\xi\in\mathcal{B}_t}\mathbb{E}_\xi\left[\|g_\xi(\boldsymbol{\theta}_t)-\nabla l(\boldsymbol{\theta}_t)\|^2\right]+2\eta^2\|\nabla l(\boldsymbol{\theta}_t)\|^2 \\
&\leq \frac{2\eta^2\sigma_{\text{SGD}}^2}{b}+\frac{9\eta}{5} [l(\boldsymbol{\theta}_t)-l(\boldsymbol{\theta}^*)],
\end{aligned}
\end{equation}
where the second step is by Lemma~\ref{lem:(a+b)2}, and the last step is by Lemma~\ref{lem:smooth}.

Substituting (\ref{eq:converge-5}) and (\ref{eq:converge-6}) into (\ref{eq:converge-4}), we have
\begin{equation}
\begin{aligned}
&\mathbb{E}_{\boldsymbol{\zeta}_t,\xi}\left[\|\boldsymbol{\theta}_{t+1}-\boldsymbol{\theta}^*\|^2\right] \\
&\leq \|\boldsymbol{\theta}_t-\boldsymbol{\theta}^*\|^2+\frac{4\eta\sigma_{\text{SGD}}^3}{\mu C}-\frac{46\eta }{25}[l(\boldsymbol{\theta}_t)-l(\boldsymbol{\theta}^*)]\\
&+ \frac{2\eta^2\sigma_{\text{SGD}}^2}{b}+\frac{9\eta}{5}[l(\boldsymbol{\theta}_t)-l(\boldsymbol{\theta}^*)] +d\eta^2\sigma_{\text{DP}}^2 \\
&\leq \|\boldsymbol{\theta}_t-\boldsymbol{\theta}^*\|^2+\frac{4\eta\sigma_{\text{SGD}}^3}{\mu C}-\frac{\eta }{25}[l(\boldsymbol{\theta}_t)-l(\boldsymbol{\theta}^*)]\\
&+ \frac{2\eta^2\sigma_{\text{SGD}}^2}{b}+d\eta^2\sigma_{\text{DP}}^2.
\end{aligned}
\end{equation}

Hence, one may obtain
\begin{equation}
\begin{aligned}
&\frac{1}{25(T+1)}\sum_{t=0}^T \mathbb{E}\left[l(\boldsymbol{\theta}_t)-l(\boldsymbol{\theta}^*)\right]\\
&\leq \frac{D^2}{\eta  (T+1)}+\frac{4\sigma_{\text{SGD}}^3}{\mu C}+\frac{2\eta\sigma_{\text{SGD}}^2}{b}+d\eta\sigma_{\text{DP}}^2.
\end{aligned}
\end{equation}
Exploiting Lemma~\ref{lem:x2x} and \ref{lem:sqrt(a+b)}, one may get
\begin{equation}
\begin{aligned}
&\frac{1}{25(T+1)}\sum_{t=0}^T \mathbb{E}\left[\sqrt{l(\boldsymbol{\theta}_t)-l(\boldsymbol{\theta}^*)}\right]\\
&\leq \sqrt{\frac{D^2}{\eta  (T+1)}+\frac{4\sigma_{\text{SGD}}^3}{\mu c}+\frac{2\eta\sigma_{\text{SGD}}^2}{b}+d\eta\sigma_{\text{DP}}^2} \\
&\leq\frac{D}{\sqrt{\eta  (T+1)}}+\frac{2\sigma_{\text{SGD}}^{1.5}}{\sqrt{\mu c}}+\frac{\sqrt{2\eta}\sigma_{\text{SGD}}}{\sqrt{b}}+\sqrt{d\eta}\sigma_{\text{DP}}.
\end{aligned}
\end{equation}
Therefore, we have
\begin{equation}
\begin{aligned}
&\frac{1}{T+1}\sum_{t=0}^T \mathbb{E}\left[\sqrt{l(\boldsymbol{\theta}_t)-l(\boldsymbol{\theta}^*)}\right]\\
&\leq \mathcal{O}\left(\frac{D}{\sqrt{\eta  T}}+\sqrt{\frac{\sigma_{\text{SGD}}^{3}}{\mu C}}+\frac{\sqrt{\eta}\sigma_{\text{SGD}}}{\sqrt{b}}+\sqrt{d\eta}\sigma_{\text{DP}}\right).
\end{aligned}
\end{equation}
Finally, summing up all the cases, one may get
\begin{equation}
\begin{aligned}
&\min_{t\in[0,T]} \mathbb{E}\left[\sqrt{l(\boldsymbol{\theta}_t)-l(\boldsymbol{\theta}^*)}\right] \\
&\leq \mathcal{O}\biggl(\frac{\sqrt{L}D^2}{\eta C T}+\frac{D}{\sqrt{\eta T}}+\min\left(\frac{L^{3/4}}{\mu^{5/4}}\sigma_{\text{SGD}},\sqrt{\frac{\sigma_{\text{SGD}}^3}{\mu C}}\right)\\
&+\frac{\sqrt{\eta}\sigma_{\text{SGD}}}{\sqrt{b}}+\frac{d\eta\sigma_{\text{DP}}^2\sqrt{L}}{C}+\sqrt{d\eta}\sigma_{\text{DP}}\biggr),
\end{aligned}
\end{equation}
which completes the proof.
\end{proof}

\subsection{Privacy-utility Trade-off for DPSGD-GC}
\label{discuss:privacy-utility-GC}

\begin{proposition}[Full statement of Proposition~\ref{coro:utility-GC}]
\label{coro:utility-GC-app}
Assuming that the conditions in Lemma~\ref{lem:converge-GC} are satisfied, for DPSGD-GC with $L$-smooth population risk, we have the following results.

(i) For $\sigma_{\textup{DP}}^2=\mathcal{O}\bigl(\frac{\alpha C^2 T}{\varepsilon nb}\bigr)$, we have the following inequality
    \begin{equation}
    \begin{aligned}
        &\min_{t\in[0,T]} \mathbb{E}\left[\|\nabla l(\boldsymbol{\theta}_t)\|\right]\\
        &\leq \mathcal{O} \biggl(\frac{1}{\eta CT}+\frac{1}{\sqrt{\eta T}}+\min \left\{\sigma_{\textup{SGD}}, \frac{\sigma_{\textup{SGD}}^2}{C}\right\}\\
        &+\sqrt{\eta L} \frac{\sigma_{\textup{SGD}}}{\sqrt{b}}+\frac{\alpha dL\eta CT}{\varepsilon n b}+\frac{\sqrt{\alpha dL\eta T}C}{\sqrt{\varepsilon nb}}\biggr).
    \end{aligned}
    \end{equation}
(ii) If we further assume the mini-batch size $b\leq \frac{n}{5}$, RDP parameter $\alpha\leq\alpha^*(\frac{b}{n},\frac{b\sqrt{\beta}\sigma_{\textup{DP}}}{2C})$, and the per-dimension Gaussian noise scale $\sigma_{\textup{DP}}>\frac{8C}{b\sqrt{\beta}}$ with a constant $\beta\in(0,1)$, and let $\sigma_{\textup{DP}}^2=\mathcal{O}\bigl(\frac{\alpha C^2 T}{\varepsilon n^2}\bigr)$, then we have the following inequality
    \begin{equation}
    \label{eq:utility-GC}
    \begin{aligned}
        &\min_{t\in[0,T]} \mathbb{E}\left[\|\nabla l(\boldsymbol{\theta}_t)\|\right]\\
        &\leq \mathcal{O} \biggl(\frac{1}{\eta CT}+\frac{1}{\sqrt{\eta T}}+\min \left\{\sigma_{\textup{SGD}}, \frac{\sigma_{\textup{SGD}}^2}{C}\right\}\\
        &+\sqrt{\eta L} \frac{\sigma_{\textup{SGD}}}{\sqrt{b}}+\frac{\alpha dL\eta CT}{\varepsilon n^2}+\frac{\sqrt{\alpha d L\eta T}C}{\sqrt{\varepsilon}n}\biggr).
    \end{aligned}
    \end{equation}
\end{proposition}

\textbf{Privacy-utility trade-off analysis.} In this subsection, we discuss the privacy-utility trade-off for DPSGD-GC. We start by Equation~(\ref{eq:utility-GC}) in Proposition~\ref{coro:utility-GC-app}, i.e.,
\begin{equation}
\begin{aligned}
    &\min_{t\in[0,T]} \mathbb{E}\left[\|\nabla l(\boldsymbol{\theta}_t)\|\right]\\
    &\leq \mathcal{O} \biggl(\frac{1}{\eta CT}+\frac{1}{\sqrt{\eta T}}+\min \bigl\{\sigma_{\textup{SGD}}, \frac{\sigma_{\textup{SGD}}^2}{C}\bigr\}\\
    &+\sqrt{\eta L} \frac{\sigma_{\textup{SGD}}}{\sqrt{b}}+\frac{\alpha dL\eta CT}{\varepsilon n^2}+\frac{\sqrt{\alpha d L\eta T}C}{\sqrt{\varepsilon}n}\biggr).
\end{aligned}
\end{equation}

We first transform $(\alpha,\varepsilon)$-RDP into the standard $(\epsilon,\delta)$-DP characterization. Let $\epsilon>0$ and $0<\delta<1$ be two constants such that $\epsilon\leq 2\log(1/\delta)$. To do the transformation, we use Lemma~\ref{lem:RDP2DP} by setting $\alpha=1+\frac{2}{\epsilon}\log(1/\delta)$ and $\varepsilon=\epsilon/2$, obtaining
\begin{equation}
\label{eq:non-square-GC}
\begin{aligned}
    &\min_{t\in[0,T]} \mathbb{E}\left[\|\nabla l(\boldsymbol{\theta}_t)\|\right]\\
    &\leq \mathcal{O} \biggl(\frac{1}{\eta CT}+\frac{1}{\sqrt{\eta T}}+\min \bigl\{\sigma_{\textup{SGD}}, \frac{\sigma_{\textup{SGD}}^2}{C}\bigr\}+\sqrt{\eta L} \frac{\sigma_{\textup{SGD}}}{\sqrt{b}}\\
    &+\frac{ dL\eta CT \log(1/\delta)}{\epsilon^2 n^2}+\frac{\sqrt{ dL\eta T \log(1/\delta)}C}{\epsilon n}\biggr).
\end{aligned}
\end{equation}
Take the square for both sides in (\ref{eq:non-square-GC}), we have
\begin{equation}
\begin{aligned}
    &\min_{t\in[0,T]}\left\{\mathbb{E}\left[\|\nabla l(\boldsymbol{\theta}_t)\|\right]\right\}^2 \\
    &\leq \mathcal{O} \biggl(\frac{1}{\eta^2 C^2T^2}+\frac{1}{\eta T}+\min \bigl\{\sigma_{\textup{SGD}}^2, \frac{\sigma_{\textup{SGD}}^4}{C^2}\bigr\}+\eta L \frac{\sigma_{\textup{SGD}}^2}{b}\\
    &+\frac{ d^2L^2\eta^2 C^2T^2 [\log(1/\delta)]^2}{\epsilon^4 n^4}+\frac{dL\eta T \log(1/\delta)C^2}{\epsilon^2n^2}\biggr).
\end{aligned}
\end{equation}
Minimizing the right-hand side terms with respect to $T$ gives $T=\Theta(\sqrt{\frac{1}{dL\log(1/\delta)}}\frac{\epsilon n}{\eta C})$. Then, we have
\begin{equation}
\begin{aligned}
    &\min_{t\in[0,T]}\left\{\mathbb{E}\left[\|\nabla  l(\boldsymbol{\theta}_t)\|\right]\right\}^2 \\
    &\leq \mathcal{O} \biggl(\frac{C\sqrt{dL\log(1/\delta)}}{\epsilon n}   
    +\frac{dL\log(1/\delta)}{\epsilon^2n^2}\\
    &+\min \left\{\sigma_{\textup{SGD}}^2, \frac{\sigma_{\textup{SGD}}^4}{C^2}\right\}+\eta L \frac{\sigma_{\textup{SGD}}^2}{b}\biggr).
\end{aligned}
\end{equation}

(1) $\frac{dL\log(1/\delta)}{\epsilon^2n^2}\geq \sigma_{\text{SGD}}^2$ (i.e., case of small stochastic noise)

In this case, we can obtain 
\begin{equation}
\begin{aligned}
    &\min_{t\in[0,T]} \left(\mathbb{E}\left[\|\nabla l(\boldsymbol{\theta}_t)\|\right]\right)^2 \\
    &\leq \mathcal{O} \biggl(\frac{C\sqrt{dL\log(1/\delta)}}{\epsilon n}   
    +\frac{dL\log(1/\delta)}{\epsilon^2n^2}+\eta L \frac{\sigma_{\textup{SGD}}^2}{b}\biggr).
\end{aligned}
\end{equation}
We choose the largest possible value for $\eta$ and $C$ to reduce the number of iterations. Also, to ensure that the utility is not compromised, we set $C=\Theta(\frac{\sqrt{dL\log(1/\delta)}}{\epsilon n})$ and $\eta=\Theta(\frac{b}{L})$. 

Consequently, we derive the following privacy-utility trade-off:
\begin{equation}
\begin{aligned}
    \min_{t\in[0,T]} \left\{\mathbb{E}\left[\|\nabla l(\boldsymbol{\theta}_t)\|\right]\right\}^2
    \leq \mathcal{O} \biggl(\frac{dL\log(1/\delta)}{\epsilon^2n^2}\biggr),
\end{aligned}
\end{equation}
with
\begin{equation}
    T=\Theta\left(\frac{\epsilon^2n^2}{bd\log(1/\delta)}\right).
\end{equation}

(2) $\frac{dL\log(1/\delta)}{\epsilon^2n^2}\leq \sigma_{\text{SGD}}^2$ (i.e., case of large stochastic noise)

In this case, we have
\begin{equation}
\begin{aligned}
    &\min_{t\in[0,T]} \left\{\mathbb{E}\left[\|\nabla l(\boldsymbol{\theta}_t)\|\right]\right\}^2 \\
    &\leq \mathcal{O} \biggl(\frac{C\sqrt{dL\log(1/\delta)}}{\epsilon n}   
    \!+\!\min \bigl\{\sigma_{\textup{SGD}}^2, \frac{\sigma_{\textup{SGD}}^4}{C^2}\bigr\}\!+\!\eta L \frac{\sigma_{\textup{SGD}}^2}{b}\biggr).
\end{aligned}
\end{equation}
Let $g(C)\stackrel{\Delta}{=} \frac{C\sqrt{dL\log(1/\delta)}}{\epsilon n}+\frac{\sigma_{\textup{SGD}}^4}{C^2}$. By setting $g^\prime(C)=0$, we can obtain $C=\Theta(\frac{\sigma_{\text{SGD}}^{\frac{4}{3}}\epsilon^{\frac{1}{3}}n^{\frac{1}{3}}}{[dL\log(1/\delta)]^{\frac{1}{6}}})$. Then, we have
\begin{equation}
    \min_{t\in[0,T]} \left\{\mathbb{E}\left[\|\nabla l(\boldsymbol{\theta}_t)\|\right]\right\}^2
    \leq \mathcal{O} \biggl(\sigma_{\text{SGD}}^{\frac{4}{3}}\left[\frac{dL\log(1/\delta)}{\epsilon^2n^2}\right]^{\frac{1}{3}}\biggr),
\end{equation}
with
\begin{equation}
    \eta=\Theta\left(\frac{b}{L\sigma_{\text{SGD}}^{2/3}}\left[\frac{dL\log(1/\delta)}{\epsilon^2n^2}\right]^{1/3}\right)
\end{equation}
and
\begin{equation}
    T=\Theta\left(\frac{\epsilon^{4/3}n^{4/3}L}{b\left[dL\log(1/\delta)\right]^{2/3}\sigma_{\text{SGD}}^{2/3}}\right).
\end{equation}

In summary, we can obtain
\begin{equation}
\begin{aligned}
    &\min_{t\in[0,T]} \left\{\mathbb{E}\left[\|\nabla l(\boldsymbol{\theta}_t)\|\right]\right\}^2 \\
    &\leq \mathcal{O} \left(\max\left\{\frac{dL\log(1/\delta)}{\epsilon^2n^2},\sigma_{\text{SGD}}^{4/3}[\frac{dL\log(1/\delta)}{\epsilon^2n^2}]^{1/3}\right\}\right).
\end{aligned}
\end{equation}

\begin{remark}
    Note that our results use a different utility metric, $\min_{t\in[0,T]} \left\{\mathbb{E}\left[\|\nabla l(\boldsymbol{\theta}_t)\|\right]\right\}^2$, instead of the more commonly used $\frac{1}{T}\sum_{t=1}^T \mathbb{E}[\|\nabla l(\boldsymbol{\theta}_t)\|^2]$, as it better facilitates our derivation and gaining insights into the analytical results.
\end{remark}

\textbf{Implications.} Prior work (see, e.g., \cite{bassily2014private}, \cite{wang2017differentially}, and \cite{zhang2017efficient}) have demonstrated that the optimal utility of 
$(\epsilon,\delta)$-DP algorithms on Lipschitz smooth population risk functions with bounded stochastic gradient assumption is $\frac{\sqrt{dL\log(1/\delta)}}{\epsilon n}$. In the case of small stochastic noise, we achieve a utility bound of $\frac{dL\log(1/\delta)}{\epsilon^2n^2}$, which is of higher order—i.e., slightly worse than the optimal utility—due to the relaxed assumption of bounded gradients and a different utility metric. When the stochastic noise scale $\sigma_{\text{SGD}}$ is large, the error introduced by the SGD variance becomes dominant and can not be effectively reduced-consistent with our intuition.

\subsection{Privacy-utility Trade-off for DPSGD-DC}
\label{discuss:privacy-utility-DC}

\begin{proposition}[Full statement of Proposition~\ref{coro:utility-DC}]
Assuming that the conditions in Theorem~\ref{thm:converge-DC} are satisfied, for DPSGD-DC with $L$-smooth and $\mu$-strongly convex population risk, we have the following results.

(i) For $\sigma_{\textup{DP}}^2=\mathcal{O}\bigl(\frac{\alpha C^2}{\varepsilon nb}\min\{T, \overline{T}\}\bigr)$ with $\overline{T}=\frac{(1+\eta L)^2nbD^2}{\eta^2C^2}$, we have the following inequality
\begin{equation}
\begin{aligned}
    &\min_{t\in[0,T]} \mathbb{E}\left[\sqrt{l(\boldsymbol{\theta}_t)-l(\boldsymbol{\theta}^*)}\right] \\
    &\leq \mathcal{O}\biggl(\frac{\sqrt{L}D^2}{\eta C T}+\frac{D}{\sqrt{\eta T}}+ \min\left\{\frac{L^{3/4}}{\mu^{5/4}}\sigma_{\textup{SGD}},\sqrt{\frac{\sigma_{\textup{SGD}}^3}{\mu C}}\right\} \\
    &+\frac{\sqrt{\eta}\sigma_{\textup{SGD}}}{\sqrt{b}}+\frac{\alpha d\eta C\sqrt{L}}{\varepsilon nb} \min\bigl\{T, \overline{T}\bigr\} \\
    &+\sqrt{\frac{\alpha d\eta}{\varepsilon nb} \min\bigl\{T, \overline{T}\bigr\} } C\biggr).
\end{aligned}
\end{equation}

(ii) If we further assume the mini-batch size $b\leq \frac{n}{5}$, RDP parameter $\alpha\leq\alpha^*(\frac{b}{n},\frac{b\sqrt{\beta}\sigma_{\textup{DP}}}{2C})$, and the per-dimension Gaussian noise scale $\sigma_{\textup{DP}}>\frac{8C}{b\sqrt{\beta}}$ with a constant $\beta\in(0,1)$, and let $\sigma_{\textup{DP}}^2=\mathcal{O}\bigl(\frac{\alpha C^2}{\varepsilon n^2}\min\{T,\overline{T}^\prime\}\bigr)$ with $\overline{T}^\prime= \frac{(1+\eta L)^2n^2D^2}{\eta^2C^2}$, then we have the following inequality
\begin{equation}
\label{eq:utility-DC}
\begin{aligned}
    &\min_{t\in[0,T]} \mathbb{E}\left[\sqrt{l(\boldsymbol{\theta}_t)-l(\boldsymbol{\theta}^*)}\right] \\
    &\leq \mathcal{O}\biggl(\frac{\sqrt{L}D^2}{\eta C T}+\frac{D}{\sqrt{\eta  T}}+ \min\left\{\frac{L^{3/4}}{\mu^{5/4}}\sigma_{\textup{SGD}},\sqrt{\frac{\sigma_{\textup{SGD}}^3}{\mu C}}\right\} \\
    &+\frac{\sqrt{\eta}\sigma_{\textup{SGD}}}{\sqrt{b}}+\frac{\alpha d\eta C\sqrt{L}}{\varepsilon n^2}\min\{T, \overline{T}^\prime\}\\
    &+\sqrt{\frac{\alpha d\eta}{\varepsilon}\min\{T,\overline{T}^\prime\}}\frac{C}{n}\biggr).
\end{aligned}
\end{equation}

\end{proposition}

We next provide the privacy-utility trade-off for DPSGD-DC in the following. We start by Equation (\ref{eq:utility-DC}) in Proposition~\ref{coro:utility-DC}, i.e.,
\begin{equation}
\begin{aligned}
    &\min_{t\in[0,T]} \mathbb{E}\left[\sqrt{l(\boldsymbol{\theta}_t)-l(\boldsymbol{\theta}^*)}\right] \\
    &\leq \mathcal{O}\biggl(\frac{\sqrt{L}D^2}{\eta C T}+\frac{D}{\sqrt{\eta  T}}+ \min\bigl\{\frac{L^{3/4}}{\mu^{5/4}}\sigma_{\textup{SGD}},\sqrt{\frac{\sigma_{\textup{SGD}}^3}{\mu C}}\bigr\} \\
    &+\frac{\sqrt{\eta}\sigma_{\textup{SGD}}}{\sqrt{b}}+\frac{\alpha d\eta C\sqrt{L}}{\varepsilon n^2}\min\{T, \overline{T}^\prime\}\\
    &+\sqrt{\frac{\alpha d\eta}{\varepsilon}\min\{T,\overline{T}^\prime\}}\frac{C}{n}\biggr).
\end{aligned}
\end{equation}

Following the transformation steps in the previous subsection, we can obtain
\begin{equation}
\begin{aligned}
    &\min_{t\in[0,T]} \left\{\mathbb{E}\left[\sqrt{l(\boldsymbol{\theta}_t)-l(\boldsymbol{\theta}^*)}\right]\right\}^2 \\
    &\leq \mathcal{O}\biggl(\frac{LD^4}{\eta^2 C^2 T^2}+\frac{D^2}{\eta  T}+ \min\bigl\{\frac{L^{3/2}}{\mu^{5/2}}\sigma_{\textup{SGD}}^2,\frac{\sigma_{\textup{SGD}}^{3}}{\mu C}\bigr\} +\frac{\eta\sigma_{\textup{SGD}}^2}{b}\\
    &+\frac{ d^2L\eta^2 C^2 [\log(1/\delta)]^2}{\epsilon^4 n^4}\min\{T^2, \overline{T}^{\prime 2}\}\\
    &+\frac{d\eta\log(1/\delta)C^2}{\epsilon^2n^2}\min\{T,\overline{T}^\prime\}\biggr),
\end{aligned}
\end{equation}
where $\overline{T}^\prime\stackrel{\Delta}{=}\frac{(1+\eta L)^2n^2D^2}{\eta^2C^2}$.

(1) $T\leq \overline{T}$

In this case, we can follow a similar way in Appendix~\ref{discuss:privacy-utility-GC} to derive the privacy-utility trade-off. First, we pick $T=\Theta(\frac{\epsilon nD}{\eta C}\sqrt{\frac{1}{d\log(1/\delta)}})$ to minimize the right hand side, resulting
\begin{equation}
\begin{aligned}
    &\min_{t\in[0,T]} \left\{\mathbb{E}\left[\sqrt{l(\boldsymbol{\theta}_t)-l(\boldsymbol{\theta}^*)}\right]\right\}^2 \\
    &\leq \mathcal{O}\biggl(\frac{CD\sqrt{d\log(1/\delta)}}{\epsilon n}+\frac{D^2dL\log(1/\delta)}{\epsilon^2n^2}\\
    &+ \min\bigl\{\frac{L^{3/2}}{\mu^{5/2}}\sigma_{\textup{SGD}}^2,\frac{\sigma_{\textup{SGD}}^{3}}{\mu C}\bigr\} +\frac{\eta\sigma_{\textup{SGD}}^2}{b}\biggr),
\end{aligned}
\end{equation}

(1.1) $\frac{D^2dL\log(1/\delta)}{\epsilon^2n^2}\geq\frac{L^{3/2}}{\mu^{5/2}}\sigma_{\textup{SGD}}^2$ (i.e., case of small stochastic noise)

In this sub-case, we pick $C=\Theta(\frac{DL\sqrt{d\log(1/\delta)}}{\epsilon n})$ and $\eta=\Theta(\frac{bL^{3/2}}{\mu^{5/2}})$, yielding 
\begin{equation}
    \min_{t\in[0,T]} \left\{\mathbb{E}\left[\sqrt{l(\boldsymbol{\theta}_t)-l(\boldsymbol{\theta}^*)}\right]\right\}^2 \leq \mathcal{O}\biggl(\frac{D^2dL\log(1/\delta)}{\epsilon^2n^2}\biggr),
\end{equation}
with
\begin{equation}
    T=\Theta\left(\frac{\epsilon^2n^2\mu^{5/2}}{bL^{3/2}\log(1/\delta)}\right).
\end{equation}

(1.2) $\frac{D^2dL\log(1/\delta)}{\epsilon^2n^2}\leq\frac{L^{3/2}}{\mu^{5/2}}\sigma_{\textup{SGD}}^2$ (i.e., case of large stochastic noise)

In this sub-case, we pick $C=\Theta(\frac{\sigma_{\text{SGD}}^{\frac{3}{2}}\epsilon^{\frac{1}{2}}n^{\frac{1}{2}}}{\mu^{\frac{1}{2}}D^{\frac{1}{2}}[dL\log(1/\delta)]^{\frac{1}{4}}})$ and $\eta=\Theta\left(\frac{bD^{\frac{1}{2}}}{\mu^{\frac{1}{2}}\sigma_{\text{SGD}}^{\frac{1}{2}}}\left[\frac{dL\log(1/\delta)}{\epsilon^2n^2}\right]^{\frac{1}{3}}\right)$. Then, we have
\begin{equation}
\begin{aligned}
    &\min_{t\in[0,T]} \left\{\mathbb{E}\left[\sqrt{l(\boldsymbol{\theta}_t)-l(\boldsymbol{\theta}^*)}\right]\right\}^2 \\
    &\leq \mathcal{O} \biggl(\frac{\sigma_{\text{SGD}}^{\frac{3}{2}}D^{\frac{1}{2}}}{\mu^{\frac{1}{2}}}\left[\frac{dL\log(1/\delta)}{\epsilon^2n^2}\right]^{\frac{1}{4}}\biggr),
\end{aligned}
\end{equation}
with
\begin{equation}
    T=\Theta\left(\frac{\epsilon^{7/6}n^{7/6}\mu D}{b\left[dL\log(1/\delta)\right]^{2/3}\sigma_{\text{SGD}}}\right).
\end{equation}

(2) $T\geq \overline{T}$

In this case, the utility bound can be expressed as
\begin{equation}
\begin{aligned}
   &\min_{t\in[0,T]} \left\{\mathbb{E}\left[\sqrt{l(\boldsymbol{\theta}_t)-l(\boldsymbol{\theta}^*)}\right]\right\}^2 \\
   &\leq \mathcal{O}\biggl(\frac{LD^4}{\eta^2 C^2 T^2}+\frac{D^2}{\eta  T}+ \min\bigl\{\frac{L^{3/2}}{\mu^{5/2}}\sigma_{\textup{SGD}}^2,\frac{\sigma_{\textup{SGD}}^{3}}{\mu C}\bigr\} \\
    &+\frac{\eta\sigma_{\textup{SGD}}^2}{b}+\frac{ d^2L(1+\eta L)^4 D^4 [\log(1/\delta)]^2}{\eta^2C^2\epsilon^4}\\
    &+\frac{d (1+\eta L)^2D^2\log(1/\delta)}{\eta \epsilon^2}\biggr),
\end{aligned}
\end{equation}
By setting $\eta=\Theta(\frac{D\sqrt{bd\log(1/\delta)}}{\epsilon\sigma_{\text{SGD}}})$ and choosing relatively large values for $C$ and $T$, we can control the utility bound to be
\begin{equation}
\begin{aligned}
    &\min_{t\in[0,T]} \left\{\mathbb{E}\left[\sqrt{l(\boldsymbol{\theta}_t)-l(\boldsymbol{\theta}^*)}\right]\right\}^2 \\
    &\leq \mathcal{O}\biggl(\sigma_{\text{SGD}}\frac{D \sqrt{d\log(1/\delta)}}{\sqrt{b}\epsilon}\biggr),
\end{aligned}
\end{equation}


In summary, we can obtain
\begin{equation}
\begin{aligned}
    &\min_{t\in[0,T]} \left\{\mathbb{E}\left[\sqrt{l(\boldsymbol{\theta}_t)-l(\boldsymbol{\theta}^*)}\right]\right\}^2 \\
    &\leq \mathcal{O} \biggl(\max\biggl\{\frac{D^2dL\log(1/\delta)}{\epsilon^2n^2},\\
    &\frac{\sigma_{\text{SGD}}^{3/2}D^{1/2}}{\mu^{1/2}}[\frac{dL\log(1/\delta)}{\epsilon^2n^2}]^{1/4},\frac{\sigma_{\text{SGD}}D \sqrt{d\log(1/\delta)}}{\sqrt{b}\epsilon}\biggr\}\biggr).
\end{aligned}
\end{equation}

\begin{remark}
    Note that our results use a different utility metric, $\min_{t\in[0,T]} \left\{\mathbb{E}\left[\sqrt{l(\boldsymbol{\theta}_t)-l(\boldsymbol{\theta}^*)}\right]\right\}^2$, instead of the more commonly used $\frac{1}{T}\sum_{t=1}^T \mathbb{E}[l(\boldsymbol{\theta}_t)-l(\boldsymbol{\theta}^*)]$, as it better facilitates our derivation and gaining insights into the analytical results.
\end{remark}

\paragraph{Implications.}

Prior work (see, e.g., \cite{bassily2014private}, \cite{wang2017differentially}, and \cite{zhang2017efficient}) have also demonstrated that the optimal utility of 
$(\epsilon,\delta)$-DP algorithms on Lipschitz smooth and strongly convex population risk functions with bounded stochastic gradient assumption is $\frac{dL\log(1/\delta)}{\mu\epsilon^2 n^2}$. In the case of small stochastic noise and small total iterations $T$, our resulting utility bound matches the theoretical bound---up to a constant---for the best attainable utility. For other cases, the error introduced by the SGD variance and domain norm becomes dominant and can not be effectively reduced-consistent with our intuition.

\section{Experimental Details}
\label{experiment-details}

\subsection{Numerical Comparison}
\label{app:num-set}
\textbf{Numerical setting for Figure~\ref{fig:compare-theoretical}.} We set the smoothness constant $L=1$, gradient clipping norm $C = 2$, noise standard deviation $\sigma_{\text{DP}} =4$, domain diameter $D = 1$, dataset size $n = 8$, batch size $b=2$, weakly convexity parameter $m=1$, privacy parameter $\alpha=1.1$, and the step size $\eta = 0.2$.

\subsection{MIA Setting}
We train a private ResNet-18 network \cite{he2016deep} as the target model using the Opacus library \cite{yousefpour2021opacus} on the standard CIFAR10 dataset \cite{krizhevsky2009learning}. According to standard MIA protocols \cite{shokri2017membership}, the training and test sets for each target and shadow model are randomly selected, equal in size, and mutually disjoint. We set the training set size to $10, 000$. While the target model's dataset does not overlap with those of the shadow models, different shadow models may partially share the same data. The shadow models are trained using the same architectures as the target model. For the attack model, we employ a two-layer multilayer perceptron (MLP) with $50$ hidden nodes and ReLU as activation functions. 

\subsection{Missing Figure for Training Loss}
\label{app:miss-fig}

\begin{figure}[ht]
\vskip 0.2in
\begin{center}
\centerline{
\includegraphics[width=0.46\textwidth]{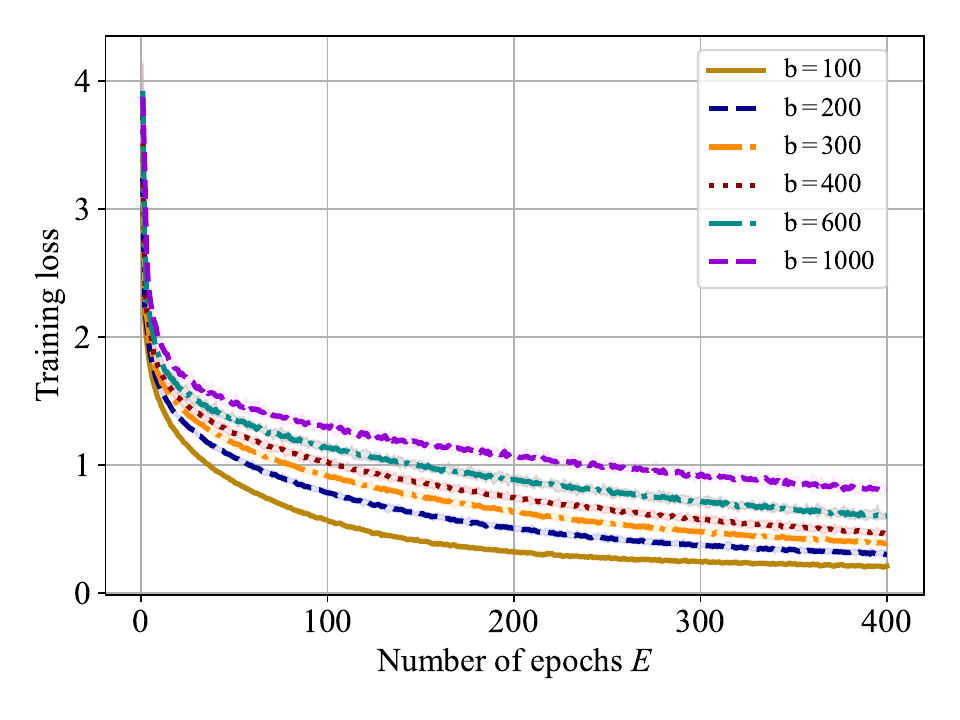}
}
\caption{The evolution of the training loss during DPSGD-GC with different batch sizes. The shaded error bars correspond to intervals covering $95$\% of the realized values, obtained from the $10$ Monte Carlo trials. Note that the utility bounds in terms of the number of epochs, $E$, can be derived by substituting $T=\lceil \frac{n}{b}\rceil E$ into our main results.}
\label{fig:loss-GC}
\end{center}
\vskip -0.2in
\end{figure}

We report the evolution of the training loss of DPSGD-GC with various batch sizes in Figure~\ref{fig:loss-GC}.

\subsection{Implementation Details}
For the target model, we use the SGD optimizer with the learning rate $\eta=0.1$, the noise level $\sigma_{\text{DP}}=0.002$, the clipping threshold $C=20$, and the confidence level $\delta=10^{-5}$. We train $10$ shadow models to simulate the behavior of the target model. For the attack classifier, we use the SGD optimizer with the initial learning rate of $0.01$, the weight decay of $5\times 10^{-4}$, and the momentum of $0.9$. The mini-batch size during training is set to be $100$. We predict the labels by selecting from the model's output the class with the highest probability. Experimental results are reported by averaging over $10$ Monte Carlo trials. Our experiments are implemented in the PyTorch \cite{paszke2017automatic} framework.

\section{Further Discussion and Limitations}

\begin{figure}[ht]
\centering
\includegraphics[width=0.46\textwidth]{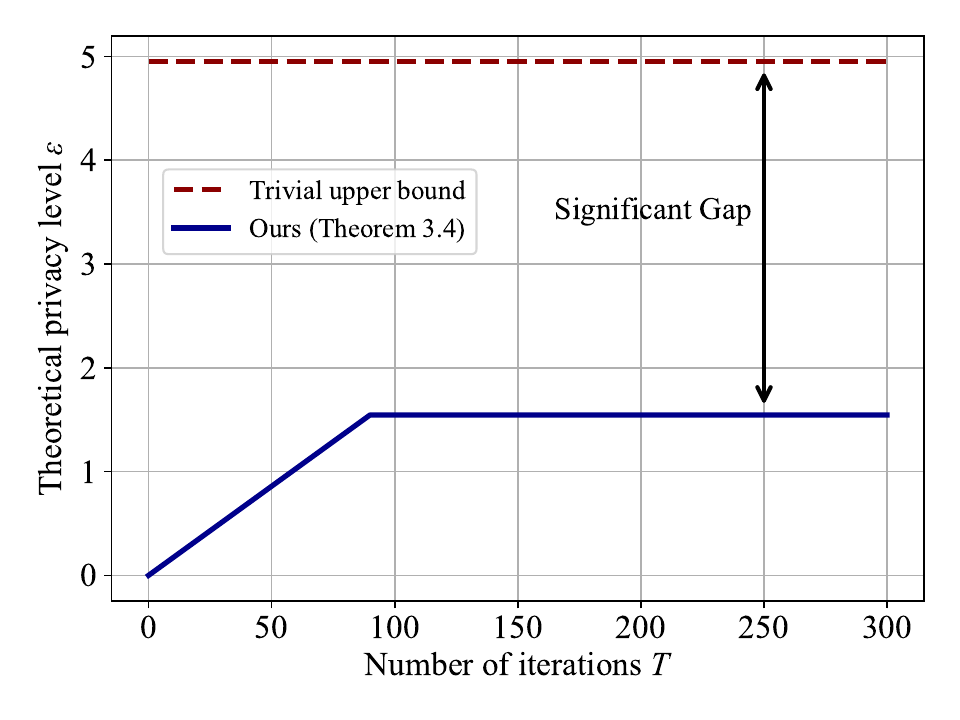}
\caption{Our tighter RDP guarantee for smooth losses during DPSGD-DC over the bounded domain, compared with the trivial bound directly utilizing the post-processing and the Gaussian mechanism property.}
\label{fig:compare-trivial}
\end{figure}

\subsection{Discussion of Bounded Domain Assumption}
\label{discuss:bounded-domain}
We provide a detailed discussion of the bounded domain assumption below. Note that a naive yet convergent bound can be obtained by combining the post-processing property (Lemma~\ref{lem:post-process}) and the standard Gaussian mechanism sensitivity analysis \cite{dwork2014algorithmic} under a bounded domain assumption. For instance, one such naive bound can be easily derived as: $\frac{2\alpha(D+\frac{\eta}{b}C)^2}{\sigma_{\text{DP}}^2}$. However, this simplistic bound severely overestimates the privacy loss, leading to overly conservative noise requirements and consequently deteriorating the privacy-utility trade-off. In this paper, we rigorously derive a tighter, non-trivial convergent privacy bound. To highlight this, we have compared our derived bound against this trivial bound in Figure~\ref{fig:compare-trivial}. For additional context, recent works such as \cite{altschuler2022privacy}, \cite{kong2024privacy}, and \cite{chien2024convergent} have also adopted the bounded domain assumption to achieve convergent guarantees.

Also, we emphasize that the bounded domain assumption is mild and practically justified. Many optimization problems naturally operate within constrained parameter domains due to intrinsic problem characteristics or practical requirements. Furthermore, some unconstrained problems may be effectively addressed by solving a sequence of constrained sub-problems, which incurs only a minor computation and privacy overhead \cite{liu2019private}. Consequently, this assumption is widely recognized as mild, practical, and standard in both optimization theory and privacy-preserving algorithmic design.

\textbf{Numerical setting for Figure~\ref{fig:compare-trivial}.} We set the smoothness constant $L=1$, gradient clipping norm $C = 2$, noise standard deviation $\sigma_{\text{DP}} =4$, domain diameter $D = 1$, dataset size $n = 16$, batch size $b=2$, privacy parameter $\alpha=1.1$, and step size $\eta = 0.2$.

\subsection{Discussion of Smoothness Assumption}
The smoothness parameter, $L$, can be estimated numerically, e.g., as the largest eigenvalue of the data Gram matrix for convex losses. While our utility analysis can be extended to broader smoothness (for example, $(L_0,L_1)$-smooth \cite{koloskova2023revisiting}), our RDP analysis requires the smooth assumption, and may generalize to Hölder continuous gradients \cite{chien2024convergent}.

\subsection{Discussion of the Selection of the Diameter $D$}
\label{discuss:select-D}
For strongly convex losses, a smaller domain diameter, $D$, improves the privacy-utility trade-off. When Theorems \ref{thm:privacy-DC} and \ref{thm:converge-DC} conditions hold, $D$ can be set as small as the optimal $\theta^{\ast}$ lies within. 

\subsection{Limitations and Future Work}
\label{discuss:limit}

\textbf{Limitations.} A key limitation of our work is that the convergent privacy bound for DPSGD-DC is derived under the assumption that the parameter domain is bounded, which may limit its direct applicability to certain practical scenarios. However, as discussed in Appendix~\ref{discuss:bounded-domain}, this assumption is mild and justifiable; in particular, some unconstrained problems can be effectively addressed by solving a sequence of constrained sub-problems. Technically, our results rely on this assumption to configure the values of auxiliary shift variables. While we provide a detailed discussion under which this is feasible and practical, deriving convergent privacy bounds without it remains a challenging open problem.

\textbf{Broader Impacts.} Our work contributes to the role that a deeper understanding of parameter projection and gradient clipping plays in DPSGD-based algorithms. This has implications for both current applications and future developments in differentially private optimization.

\textbf{Future work.} Our work implies several promising directions. First, refining the tightness of analytical results remains an open challenge. Another interesting direction is to extend this analysis to non-smooth loss functions and other optimization techniques like Adam and RMSProp, which are used notably in deep learning. Lastly, our analysis may be integrated with other optimization frameworks, such as gradient compression, distributed optimization, and federated learning.

\bibliography{my_ref}

\end{document}